%% file: main.tex
\documentclass[sigconf,anonymous=false]{acmart}

\usepackage[utf8]{inputenc}
\usepackage{amsmath,mathtools} % amssymb messes things with Bbbk
\usepackage[inline]{enumitem}
\usepackage[detect-weight=true, binary-units=true]{siunitx}
\usepackage{url,hyperref,cleveref}
\usepackage[linesnumbered]{algorithm2e}
\crefname{algocf}{algorithm}{algorithms}
\Crefname{algocf}{Algorithm}{Algorithms}
\usepackage{color,colortbl,etoolbox,xcolor} %http://tex.stackexchange.com/a/2768/22613
\robustify\bfseries
\usepackage{multirow,booktabs}
\usepackage{subcaption}
\usepackage{csquotes} %http://tex.stackexchange.com/a/294634
\usepackage[normalem]{ulem} %http://tex.stackexchange.com/a/23712
\renewcommand\vec{\boldsymbol}

\usepackage{setspace}

\usepackage[basecolor=yellow,size=0.35]{vsrs}
\input{plot-macros}

\usepackage{adjustbox}
\usetikzlibrary{shapes,arrows,positioning}
\tikzstyle{block} = [draw, fill=white, rectangle]
\tikzstyle{op} = [draw, fill=white, circle,inner sep=0.3mm]
\tikzstyle{fork} = [fill=black, circle, minimum size=1mm, inner sep=0pt, outer sep=0pt]
\tikzstyle{img} = [inner sep=0pt, outer sep=0pt]
\tikzstyle{empty} = [coordinate]
\tikzstyle{evolvable} = [text=brown]
\pgfkeys{/pgf/number format/1000 sep={\,}}

\copyrightyear{2022}
\acmYear{2022}
\setcopyright{rightsretained}
\acmConference[GECCO '22]{Genetic and Evolutionary Computation Conference}{July 9--13, 2022}{Boston, MA, USA}
\acmBooktitle{Genetic and Evolutionary Computation Conference (GECCO '22), July 9--13, 2022, Boston, MA, USA}
\acmDOI{10.1145/3512290.3528762}
\acmISBN{978-1-4503-9237-2/22/07}

\begin{document}

\title{Evolving Modular Soft Robots without Explicit Inter-Module Communication using Local Self-Attention}

\author{Federico Pigozzi}
\orcid{0000-0003-3315-6768}
\affiliation{
    \institution{DIA - University of Trieste}
    \city{Trieste} 
    \country{Italy} 
}
\email{federico.pigozzi@phd.units.it}

\author{Yujin Tang}
\orcid{0000-0003-2387-2090}
\affiliation{
    \institution{Google Brain}
    \city{Tokyo} 
    \country{Japan} 
}
\email{yujintang@google.com}

\author{Eric Medvet}
\orcid{0000-0001-5652-2113}
\affiliation{
    \institution{DIA - University of Trieste}
    \city{Trieste} 
    \country{Italy} 
}
\email{emedvet@units.it}

\author{David Ha}
\orcid{0000-0001-8765-8574}
\affiliation{
    \institution{Google Brain}
    \city{Tokyo} 
    \country{Japan} 
}
\email{hadavid@google.com}

\begin{abstract} %max 200 words (currently ~195)
Modularity in robotics holds great potential.
In principle, modular robots can be disassembled and reassembled in different robots, and possibly perform new tasks.
Nevertheless, actually exploiting modularity is yet an unsolved problem: controllers usually rely on inter-module communication, a practical requirement that makes modules not perfectly interchangeable and thus limits their flexibility.
Here, we focus on Voxel-based Soft Robots (VSRs), aggregations of mechanically identical elastic blocks.
We use the same neural controller inside each voxel, but without \emph{any} inter-voxel communication, hence enabling ideal conditions for modularity: modules are all equal and interchangeable.
We optimize the parameters of the neural controller---shared among the voxels---by evolutionary computation.
Crucially, we use a local \emph{self-attention} mechanism inside the controller to overcome the absence of inter-module communication channels, thus enabling our robots to truly be driven by the collective intelligence of their modules.
We show experimentally that the evolved robots are effective in the task of locomotion: thanks to self-attention, instances of the same controller embodied in the same robot can focus on different inputs.
We also find that the evolved controllers generalize to unseen morphologies, after a short fine-tuning, suggesting that an inductive bias related to the task arises from true modularity.\footnote{Videos of our results are available at \url{https://softrobots.github.io/}.}
\end{abstract}



%\ccsdesc[300]{Computing methodologies~Continuous space search}
%\ccsdesc[500]{Computer systems organization~Evolutionary robotics}

\begin{teaserfigure}
    \begin{tikzpicture}
        \node at (0,0) (title1) {Self-attention controller};
        \node at (0,-0.7) (input) {inputs};
        \node[fork] at (0,-1.2) (inputfork) {};
        \node[block] at (1,-1.7) (attentionmodule) {attention module};
        \node[empty] at (-1,-1.7) (x) {};
        \node[right] at (1,-2.2) (a) {$\vec{A}$};
        \node[op] at (0,-2.3) (prod) {$\times$};
        \node[block] at (0,-3) (downstreammodule) {downstream module};
        \node at (0,-3.7) (output) {output};
        \draw[->] (input) -- (inputfork) -| (attentionmodule);
        \draw[->] (inputfork) -| (x) |- (prod);
        \draw[->] (attentionmodule) |- (prod);
        \draw[->] (prod) -- (downstreammodule);
        \draw[->] (downstreammodule) -- (output);
        \node at (6.25,0) (title2) {Self-attention visualization};
        \node[img] at (3.5,-3) (patch) {\includegraphics[height=2cm]{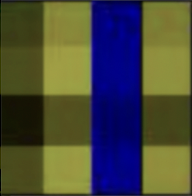}};
        \draw [dashed] (a.east) ++(0,0.5mm) -- (patch.north west);
        \draw [dashed] (a.east) ++(0,-0.5mm) -- (patch.south west);
        \node[img] at (7.5,-2.35) (robot) {\includegraphics[height=3.3cm]{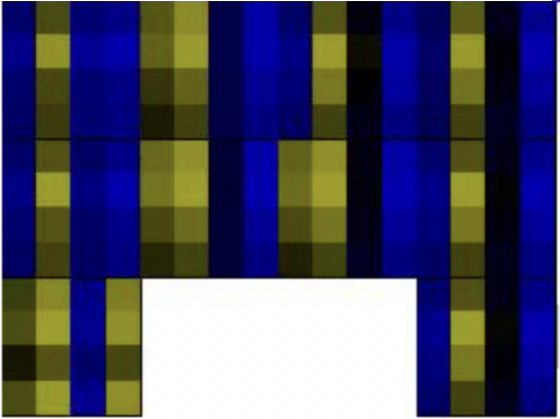}};
        \draw [dashed] (patch.south east) -- (robot.south west);
        \draw [dashed] (robot.south west) ++(0,1.1cm) -- (patch.north east);
        \node at (13.5, 0) (title3) {Generalization};
        \node[img] at (13, -1.15) (vsr) {\includegraphics[scale=0.4]{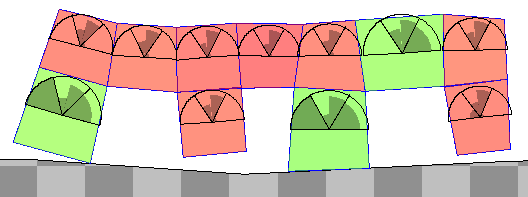}};
        \node[img] at (13, -3.15) (vsrlarge) {\includegraphics[scale=0.5]{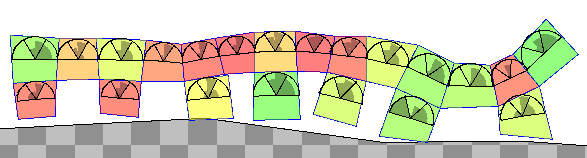}};
        \draw [->] (vsr) -- (vsrlarge);
    \end{tikzpicture}
    \caption{
        Overview of the proposed approach.
        We use the same neural controller (left picture) inside each voxel, with shared parameters.
        The middle picture is a biped with the attention matrices of the different voxels.
        Each controller uses self-attention to compute importance scores ($\vec{A}$) among the inputs sensed by its voxel.
        We also find evolved controllers to generalize to unseen morphologies (right picture; color represents the ratio between the voxel current area and its rest area: red stands for contraction, green for expansion, yellow for no change).
    }
    \vskip 0.10 in
    \label{fig:intro-picture}
\end{teaserfigure}
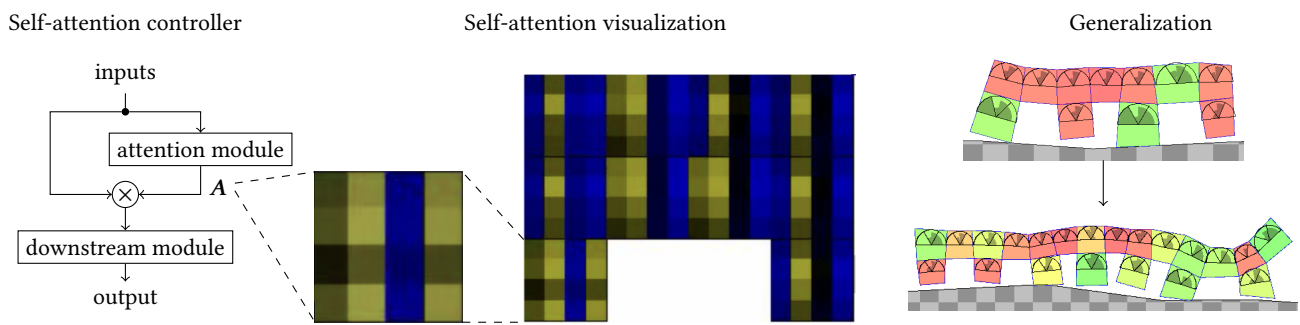

\maketitle

\section{Introduction}
\label{sec:intro}
% intro to list promise of modular robots, put VSRs in perspective of modular robots, and define acronyms
Autonomous robotic ecosystems~\cite{hale2020hardware,buchanan2020bootstrapping} promise to accomplish cooperatives of robots working toward a common, human-aligned goal, ideally for tasks that are either too dangerous (e.g., rescue missions) or unfeasible for humans (e.g., digesting pollutants).
In this setting, modularity plays a crucial role: unlike rigid, fully-integrated robots, modular robots rely on a large number of often identical independent modules that collectively complete an objective.
As such, they have the flexibility to adapt to a wide range of morphologies and environments~\cite{kriegman2021scale}, possess fault tolerant properties~\cite{christensen2013distributed}, and can even self-assemble into novel morphologies that have not been envisioned by their original designers~\cite{pathak2019learning}.
Indeed, modularity is a gift of nature to living organisms~\cite{fodor1983modularity}, and a product of evolution.

Modularity is also ubiquitous in the field of artificial intelligence: it appears in Graph Neural Networks (GNNs)~\cite{scarselli2008graph}, Cellular Automata (CA)~\cite{mordvintsev2020growing}, and multi-agent systems~\cite{wong2021multiagent}.
In evolutionary robotics, Voxel-based Soft Robots (VSRs)~\cite{hiller2012automatic} are simulated aggregations of mechanically identical elastic blocks: as such, they have emerged as as a relevant formalism to model state-of-the-art robotic systems, e.g., soft robotics~\cite{rus2015design}.
VSRs are appealing for investigating questions related to Artificial Life (AL)~\cite{kriegman2018morphological} and designing living organisms that evolve in vivo~\cite{kriegman2020scalable}, bridging the sim-to-real gap~\cite{blackiston2021cellular}.
In doing so, researchers mostly rely on Evolutionary Algorithms (EAs)~\cite{de2006evolutionary}, a family of algorithms that successfully married with the field of robotics~\cite{sims1994evolving,nolfi2000evolutionary,tang2020learning,evojax2022}.

Yet, in the path toward autonomous robotic ecosystems, full exploitation of modularity is still a roadblock~\cite{yim2007modular}, as the flexibility of modular robots depends on how the modules are ``wired'' together.
Minimizing the need for communication among modules would permit to disassemble robots and to reassemble them differently, to cope with different tasks.
Early controllers for VSRs relied on a single centralized neural network~\cite{talamini2019evolutionary} or fixed, rigid inter-module communication between voxels~\cite{medvet2020evolution,horibe2021regenerating}, effectively making their design closer to a single, large neural controller with shared weights.
This rigid Message-Passing (MP) mechanism, while practical, makes the modules not perfectly interchangeable, thus limiting their flexibility.
This leads to the question of whether VSRs will work \emph{at all} if we remove \emph{all} MP channels between modules.
Any communication will thus take place through physical interactions in the environment initiated by each module, while simultaneously performing a task collectively: an outmost instance of morphological computation~\cite{hauser2011towards} (i.e., the ``brain'' offloading computation to the ``body'').

In this work, we explore the properties of such modular robots without MP channels.
We use the same neural controller inside each voxel, but without \emph{any} inter-voxel communication, hence realizing the ideal conditions for modularity: modules are all equal and interchangeable, enabling our robots to truly be driven by the collective intelligence of their modules.
We leverage EAs to optimize the controller parameters, and experimentally test whether the resulting VSRs are effective in a locomotion task on hilly terrains.
We initially find the lack of inter-module communication to be too severe of a handicap for VSRs that rely on traditional fully-connected neural network controllers.
However, we find that a local self-attention mechanism---a form of adaptive weights---achieves superior performance since instances of the same controller embodied in the same robot can focus on different inputs.
See \Cref{fig:intro-picture} for an overview of the proposed approach.
Further, we also find that we can generalize to unseen morphologies after a short fine-tuning, suggesting that an inductive bias related to the task arises from true modularity.
Through these findings, we envision this work to position itself as a stepping stone in the road toward autonomous robotic ecosystems.

\section{Related work}
\label{sec:related}
% Talk about modular robots, and their benefits.
Modularity allows robotic systems to present various kinematic configurations beyond what a fixed architecture can, and such robots are usually optimal to solve many robotic tasks~\cite{siciliano2008springer,eiben2021real}.
Modularity in robotic systems takes the form of fabricating physical parts that are interchangeable for a single robot, or designing independent robots that participate in a common task adaptively~\cite{faina2021evolving}.
Due to the flexibility, versatility, and robustness to changing environmental conditions~\cite{yim2007modular}, we are witnessing an increasing number of works on modular robots~\cite{howison2020reality}, including soft ones~\cite{sui2020automatic}.
For example, \citet{kamimura2003automatic} proposed a method to automatically generate locomotion patterns for an arbitrary configuration, and \citet{gross2006object} tackled object manipulation and transportation tasks using self-reconfigurable swarm-bots.
The possibilities of such systems are encouraging, but the self-organizing and adaptive properties are even more inspiring.
In a life-long gait learning task, \citet{christensen2013distributed} showed that, given a body configuration, a modular system can automatically figure out the best gaits, while presenting morphology independence and fault tolerance.
Taking a step further, \citet{pathak2019learning} demonstrated that a modular robotic system can generalize to unseen morphologies and tasks.
Finally, \citet{pigozzi2022embodied} partitioned modular robots into independent units on the basis of self-organizing neural controllers.

% Relate ML to modular robots.
The concept of modularity appears also in the AL and machine learning communities, most noticeably in the area of neuroevolution and CA~\cite{ha2021collective}.
Indeed, we are witnessing a surge in works that incorporate these ideas into modular robots.
While most robotic works are controller-focused, AL and ML researchers also explored the co-optimization of configuration and controller~\cite{ha2019reinforcement,lan2021learning,bhatia2021evolution}.
For instance, \citet{cheney2014unshackling} evolved soft robots with multiple materials through a generative encoding.
Inspired by multi-cellular systems, \citet{joachimczak2016metamrphosis} evolved soft-bodied animats in both aquatic and terrestrial environments, showcasing the concept of metamorphosis in simulation.
In another soft robot simulation work, partially damaged robots regenerated their original morphology through local cell interactions in a neural CA system~\cite{horibe2021regenerating}.
Moreover, \citet{sudhakaran2021growing} grew complex functional entities in Minecraft through a neural CA-based morphogenetic process.
Finally, there exists a whole body of literature on the evolution of virtual creatures with artificial gene regulatory networks~\cite{cussat2019artificial}.

Intuitively, the performance of a modularized system depends on the communication pattern, one therefore naturally wonders if there exist better MP graphs.
Recent work suggests that GNNs possess self-organizing properties and are capable of learning rules for established CA systems~\cite{grattarola2021learning}, which hints at learning novel inter-module communication patterns.
On the other hand, the very existence of inter-module MP prevents modular robots from being robust and truly interchangeable.
Researchers thus set their eyes on the other end of the spectrum and explored the possibility of creating modular robots with more localized communication~\cite{owaki2021tegotae} or even without inter-module communications~\cite{martius2013information,kalat2018decentralized,queralta2019communication}.
In this work, we evolve simulated VSRs for locomotion tasks, and demonstrate that it is possible to evolve a shared controller for each voxel that excludes communication with others.

% Related work? https://twitter.com/hardmaru/status/1239379852685242368

% embodied cognition direction. (discuss systems that work without a brain, that might apply to VSRs).

% communication-free swarm robotics
% - Kalat, Shadi Tasdighi, Siamak G. Faal, and Cagdas D. Onal. "A decentralized, communication-free force distribution method with application to collective object manipulation." Journal of Dynamic Systems, Measurement, and Control 140.9 (2018): 091012.
% - Queralta, J. Pena, et al. "Communication-free and index-free distributed formation control algorithm for multi-robot systems." Procedia Computer Science 151 (2019): 431-438.

\section{Methods}
\label{sec:methods}

\subsection{Background on Voxel-based Soft Robots}
\label{sec:vsrs}
Voxel-based Soft Robots (VSRs) are a kind of modular soft robots composed as aggregations of elastic cubic blocks (\emph{voxels}).
Each voxel contracts or expands its volume; it is the overall symphony of volume changes that allows for the emergence of the high-level behavior of the robot.
\citet{hiller2012automatic} first formalized VSRs and proposed a fabrication method.
In this work, we consider a 2-D variant of simulated (in discrete time and continuous space) VSRs~\cite{medvet20202d}, where cubes become squares.
While disregarding one dimension certainly makes these simulated VSRs less realistic, it also eases the optimization of VSR design and facilitates research on their control~\cite{bhatia2021evolution}, thanks to the smaller search space.
However, the representations and the algorithms adopted in this paper are easily portable to the 3-D setting, and so are the considerations on the self-attention controller.

In the following, we outline the characteristics of VSRs relevant to this study, and refer the reader to~\cite{medvet20202d} for more details.
A VSR is completely defined by its \emph{morphology} (i.e., the body) and its \emph{controller} (i.e., the brain), that we detail in the following two subsections.

\subsubsection{VSR morphology}
\label{sec:morphology}
The morphology of a VSR describes how the voxels, i.e., elastic squares, are arranged in a grid topology.
Each voxel is modeled as the assembly of spring-damper systems, masses, and distance constraints~\cite{medvet20202d}.
Each voxel is rigidly connected to its four adjacent voxels (if present).

Over time, voxels change their area according to
\begin{enumerate*}[label=(\alph*)]%
    \item external forces acting on the voxel (e.g., other bodies, including other voxels and the ground) and
    \item an actuation signal computed by the controller.
\end{enumerate*}
The latter produces a contraction/expansion force that is modeled in the simulation as an instantaneous change in the resting length of the spring-damper systems of the voxel.
The length change is linearly dependent on an actuation value residing in $[-1,+1]$, $-1$ being the greatest possible expansion and $+1$ being the greatest possible contraction.

Each voxel receives, at every time step of the simulation, the readings $\vec{s} \in \mathbb{R}^4$ of four sensors embedded in the voxels.
Area sensors sense the ratio between the current area of the voxel and its resting area.
Touch sensors sense whether the voxel is touching another body different from the robot itself (e.g., the ground) or not, and output a value of $1$ or $0$, respectively.
Velocity sensors sense the speed of the center of mass of the voxel along the $x$- and $y$-directions.
We normalize all sensor readings into $[0,1]^4$ by soft normalization.
After normalization, to simulate real-world sensor noise, we perturb every sensor reading with additive Gaussian noise with mean $0$ and variance $\sigma_\text{noise}^2$.
We set $\sigma_\text{noise}=0.01$.

\subsubsection{Controller}
\label{sec:controller}
We consider the distributed controller proposed in~\cite{medvet2020evolution}, consisting in a number of fully-connected, feed-forward Artificial Neural Networks (ANNs), one for every voxel.
In particular, we adopt the ``homogeneous'' variant presented in~\cite{medvet2021biodiversity}, where the ANNs share the same parameters: \citet{medvet2021biodiversity} proved that such homogeneous representation is comparable to one where parameters are different for every ANN, with the additional benefit of a more compact search space, similarly to what happens in most multi-agent reinforcement learning systems~\cite{wong2021multiagent}.
Moreover, parameter sharing makes the controller agnostic with respect to the morphology, putting ourselves on a vantage point to test generalization to unseen morphologies.
With respect to the model of~\cite{medvet2020evolution,medvet2021biodiversity}, we here drop MP among voxels, as the focus of this work is on minimizing (and, possibly, dispensing with) inter-module communication in soft robots: we hence enable interchangeability of modules and thus the full exploitation of modularity.
Moreover, we perform actuation every $k_\text{act}$ steps, rather than at every time step as in~\cite{medvet2020evolution,medvet2021biodiversity}, to prevent VSRs from exploiting the emergence of high-frequency dynamics.

Every ANN takes as input the local sensor readings $\vec{s} \in \mathbb{R}^4$ and outputs the local actuation value $a \in \mathbb{R}$.
We use a one-hot encoding of $\vec{s} \in \mathbb{R}^4$ to let the voxels know where they are in the body: the actual input of the ANN is hence built as follows.
Let $n$ be the number of voxels of a given VSR; for the $i$-th voxel, let $\vec{h}_i \in \mathbb{R}^n$ be a one-hot vector that is $1$ at the $i$-th entry and $0$ everywhere else.
We encode $\vec{s}$ for each $i$-th voxel as $\vec{X}=\vec{s}\vec{h}_i^T \in \mathbb{R}^{4 \times n}$.
$\vec{X}$ is then a matrix that is equal to $\vec{s}$ at the $i$-th column and $0$ otherwise. 
In this way, a voxel can distinguish itself in the morphology.
We apply this same pre-processing to every model considered in this paper.

We remark that one-hot encoding does not invalidate the claim that voxels are all identical: in practice, it just requires the ``operator'' (i.e., the robot assembler) to set the position of each voxel ``in'' the voxel controller itself, i.e., to do the proper configuration.
Regardless of the ANN architecture, there is a unique vector $\vec{\theta} \in \mathbb{R}^p$ of parameters that specifies every ANN in the VSR.
Thus, we optimize a VSR for a given task by optimizing the parameters $\vec{\theta}$.

% potentially other forms of adaptive weights
\subsection{Self-attention}
\label{sec:attention}
Attention can be seen as an ``adaptive weights''~\cite{ferigo2021evolving} mechanism that computes importance scores for the inputs.
Attention mechanisms were first introduced in the context of machine translation~\cite{bahdanau2014neural,luong2015effective} to capture relationships in temporal sequences of data, and have thus prospered in natural language processing~\cite{devlin2018bert,galassi2020attention}.
Attention mechanisms have achieved state-of-the-art performance in domains (e.g., computer vision~\cite{khan2021transformers}) where data are not temporal but spatial~\cite{dosovitskiy2020image,wu2020visual} or even sets~\cite{lee2019set,tang2021sensory}, also considering robotic settings~\cite{choi2017multi,zambaldi2018deep,tang2020neuroevolution}.
There is indeed evidence that the nervous system modulates attention on every sensory channel~\cite{driver2001selective}.
As a result, being attention an adaptive weights mechanism, we argue it could supplant inter-module communication by computing importance scores that are tailored to the specific voxel, while being local (i.e., attending only to voxel-specific information) and shared (i.e., same parameters for every voxel).

Let $\vec{X} \in \mathbb{R}^{r \times u}$ be a sequence of $u$ inputs of dimension $r$.
In its general formulation, an attention module computes an \emph{attention matrix} $\vec{A} \in \mathbb{R}^{r \times r}$, to get weighted inputs:
\begin{equation}
    \vec{Y} = \vec{A}\vec{X}
    \label{eq:attention}
\end{equation}
to be fed to a downstream module $f(\vec{Y})=z \in \mathbb{R}$ for some desired output $z$.

While a great variety of attention mechanisms do exist in the literature~\cite{chaudhari2021attentive}, we resort on self-attention~\cite{vaswani2017attention}.
Self-attention is a generalized form of attention, and has already shown to achieve state-of-the-art results on continuous control tasks that exploit data other than temporal~\cite{tang2020neuroevolution}.
Self-attention computes:
\begin{equation}
    \vec{A} = \sigma\left(\frac{1}{\sqrt{d}}\vec{Q}\vec{K}^T\right)
    \label{eq:self-attention}
\end{equation}
where $\sigma(\cdot)$ is a non-linear function that constrains $\vec{A}$ to be in a given range (e.g., $\tanh$), $\vec{Q},\vec{K} \in \mathbb{R}^{r \times d}$ (known as the Query and Key matrices) are the output of linear transformations of the form:
\begin{align}
    \vec{Q} &= \vec{X} \vec{W}_q + \vec{b}_q \\
    \vec{K} &= \vec{X} \vec{W}_k + \vec{b}_k
\end{align}
where $\vec{W}_q \in \mathbb{R}^{u \times d}$, $\vec{W}_k \in \mathbb{R}^{u \times d}$ are weight matrices, $\vec{b}_q \in \mathbb{R}^{d}$, $\vec{b}_k \in \mathbb{R}^{d}$ are bias vectors, and $+$ denotes the matrix-vector addition (the vector is added to each row of the matrix).
We take the dot product between $\vec{Q}$ and $\vec{K}$ to compute compatibility between two different representations of the inputs.
The division in \Cref{eq:self-attention} appears because the dot product grows with the operands dimensions.

\subsubsection{Self-attention in VSR controller}
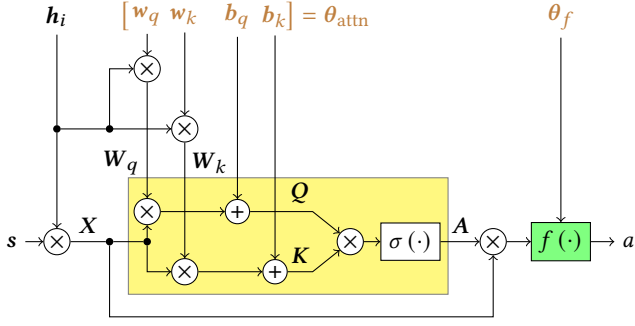
\begin{figure}
    \centering
    \begin{tikzpicture}
        \draw[fill=yellow,opacity=0.5] (2.25,-0.7) rectangle (6.5,0.85);

        \node (s) at (0.7,0) {$\vec{s}$};
        \node[op] (prod1) at (1.3,0) {$\times$};
        \node (h) at (1.3,3) {$\vec{h}_i$};
        \node[fork] (hfork) at (1.3,1.5) {};
        \node[fork] (xfork) at (2,0) {};
        \node[fork] (xfork2) at (2.5,0) {};
        \node[fork] (hfork2) at (2,1.5) {};
        \node[evolvable] (wq) at (2.5,3) {$\vec{w}_q$};
        \node[evolvable] (wk) at (3,3) {$\vec{w}_k$};
        \node[op] (prod2q) at (2.5,0.4) {$\times$};
        \node[op] (prod2k) at (3,-0.4) {$\times$};
        \node[op] (prod3q) at (2.5,2.3) {$\times$};
        \node[op] (prod3k) at (3,1.5) {$\times$};
        \node[op] (addq) at (3.7,0.4) {$+$};
        \node[op] (addk) at (4.2,-0.4) {$+$};
        \node[empty] (q) at (4.7,0.4) {};
        \node[empty] (k) at (4.7,-0.4) {};
        \node[op] (prod4) at (5.2,0) {$\times$};
        \node[block] (sigma) at (6,0) {$\sigma\left( \cdot \right)$};
        \node[evolvable] (bq) at (3.7,3) {$\vec{b}_q$};
        \node[evolvable] (bk) at (4.2,3) {$\vec{b}_k$};
        \node[op] (prod5) at (7.1,0) {$\times$};
        \node[empty] (x) at (5,-1) {};
        \node[block,fill=green,fill opacity=0.5,text opacity=1] (f) at (8,0) {$f\left( \cdot \right)$};
        \node[evolvable] (thetaf) at (8,3) {$\vec{\theta}_f$};
        \node (a) at (8.9,0) {$a$};
        \node[evolvable] at (2.2,3) {$\big [$};
        \node[evolvable] at (4.9,3) {$\big ]=\vec{\theta}_\text{attn}$};

        \draw[->] (s) -- (prod1);
        \draw[->] (h) -- (hfork) -- (prod1);
        \draw[->] (prod1) -- node[above] {$\vec{X}$} (xfork) -- (xfork2) -- (prod2q);
        \draw[->] (xfork2) |- (prod2k);
        \draw[->] (hfork) -- (hfork2) |- (prod3k);
        \draw[->] (hfork2) |- (prod3q);
        \draw[->] (prod3q) -- node[left,yshift=-3mm] {$\vec{W}_q$} (prod2q);
        \draw[->] (prod3k) -- node[right,yshift=5mm] {$\vec{W}_k$} (prod2k);
        \draw[->] (prod2q) -- (addq);
        \draw[->] (prod2k) -- (addk);
        \draw[->] (wq) -- (prod3q);
        \draw[->] (wk) -- (prod3k);
        \draw[->] (addq) -- node[above,xshift=2.5mm] {$\vec{Q}$} (q) -- (prod4);
        \draw[->] (addk) -- node[above] {$\vec{K}$} (k) -- (prod4);
        \draw[->] (bq) -- (addq);
        \draw[->] (bk) -- (addk);
        \draw[->] (prod4) -- (sigma);
        \draw[->] (sigma) -- node[above] {$\vec{A}$} (prod5);
        \draw[->] (xfork) |- (x) -| (prod5);
        \draw[->] (prod5) -- (f);
        \draw[->] (thetaf) -- (f);
        \draw[->] (f) -- (a);
        
    \end{tikzpicture}
    \caption{
        The architecture of our self-attention controller.
        Evolvable parameters are shown in brown, the attention module is shown in yellow, the downstream module is shown in green.
        Parameters are the same for every voxel.
    }
    \label{fig:self-attention}
\end{figure}

We use the one-hot encoded sensor reading as input $\vec{X}$: then, $r=4$ and $u=n$.
The attention matrix is $\vec{A} \in \mathbb{R}^{4 \times 4}$: the attention is on the sensor readings and $\vec{A}$ dimension does not depend on the robot morphology.
For self-attention to focus only on local information, we set on each $i$-th voxel $\vec{W}_q = \vec{h}_i \vec{w}_q^T$ and $\vec{W}_k = \vec{h}_i \vec{w}_k^T$, with $\vec{h}_i$ defined as in \Cref{sec:controller}: in this way, we extract the column of $\vec{X}$ corresponding to the voxel.
$\vec{w}_q,\vec{w}_k \in \mathbb{R}^d$ are evolvable vectors of parameters and are the same for all the voxels.
We summarize the building blocks of our architecture in \Cref{fig:self-attention}.

After preliminary experiments, we set $\sigma(\cdot)$ to be $\tanh$, $d=8$, and $f$ to be a Multi Layer Perceptron (MLP) with no hidden layers and $\tanh$ activation function (to ensure the output lies in $[-1,+1]$).
Moreover, given that we use $\tanh$ as non-linearity in \Cref{eq:self-attention}, the entries of $\vec{A}$ lie in $[-1,+1]$, $+1$ being the highest compatibility between two inputs, $-1$ the least, and $0$ no compatibility.
Finally, our model differs from the original formulation of self-attention~\cite{vaswani2017attention} in that we set Values to be the identity function, since, after preliminary experiments, we found them to be unnecessary. 

In a self-attention model of this form, we optimize the parameters $\vec{\theta}=   \left[\vec{\theta}_\text{attn} \; \vec{\theta}_f\right]$, where the attention module parameters $\vec{\theta}_\text{attn}$ are the concatenation of $\vec{w}_q,\vec{w}_k,\vec{b}_q,\vec{b}_k$, and the downstream module parameters $\vec{\theta}_f$ are the weights and biases of the downstream MLP.
Then, $\vec{\theta}$ is the genotype of our EA, that we detail in the next subsection.

\subsection{Evolutionary algorithm}
\label{sec:ea}
We perform optimization by means of Evolutionary Computation (EC); in particular, we resort to a Genetic Algorithm (GA)~\cite{de2006evolutionary}.
Indeed, \citet{risi2019deep} used GAs to effectively evolve complex neural architectures consisting of heterogeneous modules, and \citet{such2017deep} proved GAs to be competitive with state-of-the-art reinforcement learning algorithms.

Our GA iteratively evolves a fixed-size population of $n_\text{pop}$ individuals according to a $\mu + \lambda$ generational model, i.e., with overlapping.
We initialize individuals at the very first iteration by uniformly sampling the interval $[-1,+1]^p$.
Then, at each iteration, we select parents with tournament selection of size $n_\text{tour}$ and build the offspring applying Gaussian mutation, with probability $p_\text{mut}$, or extended geometric crossover, with probability $1-p_\text{mut}$.
For mutation, given a parent $\vec{\theta} \in \mathbb{R}^p$, we add Gaussian noise to get $\vec{\theta}' = \vec{\theta} + \sigma_\text{mut}\vec{\epsilon}$, with $\vec{\epsilon} \sim \mathcal{N}(0,\vec{I})$ and $\vec{I}$ being the diagonal matrix of size $p \times p$.
For extended geometric crossover, given two parents $\vec{\theta}_1,\vec{\theta}_2 \in \mathbb{R}^p$, the new individual is born as $\vec{\theta}' = \vec{\theta}_1 + \vec{\alpha}(\vec{\theta}_2-\vec{\theta}_1) + \sigma'_\text{mut}\vec{\epsilon}$, where each element $\alpha_i$ of $\vec{\alpha}$ is chosen randomly with uniform probability in $[-0.5,1.5]$ and $\vec{\epsilon} \sim \mathcal{N}(0,\vec{I})$.
Then, we merge offspring and parents and retain only the best half of individuals, that will constitute the population at the next iterations.
Evolution terminates after $n_\text{evals}$ fitness evaluations have been computed.

After preliminary experiments and exploiting our previous knowledge, we set $n_\text{pop}=100$, $n_\text{tour}=5$, $\sigma_\text{mut}=0.35$, $\sigma_\text{mut}'=0.1$, $p_\text{mut}=0.2$, and, unless otherwise specified, $n_\text{evals}=\num{30000}$.

\section{Experiments}
\label{sec:experiments}
Our goal is to answer the following questions with an experimental analysis:

\begin{enumerate}[label=RQ\arabic*,leftmargin=2.75\parindent]
  \item \label{item:rq-1} Are VSRs evolved with self-attention effective at solving a locomotion task? If so, are they robust to environmental changes?
  \item \label{item:rq-2} Why does self-attention work?
\end{enumerate}

We evaluate our method on two different VSR shapes, namely a $4 \times 3$ rectangle with a $2 \times 1$ rectangle of missing voxels at the bottom-center, that we call \emph{biped} \vsr[1mm]{4}{3}{1111-1111-1001}, and a $7 \times 2$ rectangle with empty voxels at the odd $x$ positions in the bottom row, that we call \emph{comb} \vsr[1mm]{7}{2}{1111111-1010101}.
The dimension of $\vec{X}$ is hence $4 \times 10$ for the biped shape and $4 \times 11$ for the comb shape (the second operand in the multiplication being the number of voxels); as a result, the self attention controller has $\left|\vec{\theta}\right|=73$ parameters for biped ($\left|\vec{\theta}_f\right|=41$ pertaining to the downstream MLP and $\left|\vec{\theta}_\text{attn}\right|=32$ for the attention module) and $77$ for comb ($\left|\vec{\theta}_f\right|=45$ and $\left|\vec{\theta}_\text{attn}\right|=32$).

For all the experiments, we considered the task of locomotion. 
The goal is to travel as fast as possible on a terrain along the positive $x$ direction, in a fixed amount of simulated time $t_\text{final}$.
We use as fitness the average velocity $\overline{v}_x$ of the center of mass of the VSR during the simulation.
We set $t_\text{final}=\SI{30}{\second}$.
Locomotion is indeed a classic task in evolutionary robotics~\cite{nolfi2000evolutionary}, but usually consists in running along a flat surface.
We hereon consider a hilly terrain, consisting of bumps that are randomly procedurally generated with an average height of $\SI{1}{\meter}$ and an average distance of $\SI{10}{\meter}$.
We randomize the seed for the procedural generation at every fitness evaluation and re-evaluate individuals retained from the previous iteration, so that evolution does not unfairly favor individuals with an ``easy'' terrain and to make adaptation more challenging.

We implemented the experimental setup in the \text{Java} programming language, relying on \text{JGEA}\footnote{\url{https://github.com/ericmedvet/jgea}.} for the evolutionary optimization and \text{2D-VSR-Sim}~\cite{medvet20202d} for the simulation of VSRs.
For the latter, we set $\Delta t = \SI[parse-numbers=false]{\frac{1}{60}}{\second}$ for the time step, and all other parameters to default values (as a result, all voxels share the same mechanical properties).
After preliminary experiments, we set $k_\text{act}=20$ (i.e., one actuation every $\approx \SI{0.33}{\second}$).
We made the code publicly available at \url{https://github.com/pigozzif/AttentionVSRs}.

For each experiment, we performed $5$ evolutionary runs by varying the random seed for the EA.
We remark that, for a given VSR and terrain, the simulations are instead deterministic.
We carried out all statistical tests with the Mann-Whitney U rank test for independent samples.

\subsection{Results}
\label{sec:results}

\subsubsection{\ref{item:rq-1}: effectiveness and robustness with self-attention}
\label{sec:rq-1}
In order to verify the effectiveness of evolved VSRs equipped with our self-attention model, we measure their performance in two different cases: in the same conditions they were evaluated during the evolution and in slightly different environmental conditions aimed at testing an individual generalization abilities.
In both cases, we use $\overline{v}_x$ as performance index: in the former, it is the value of the fitness function itself, while in the latter it is the average over $10$ unseen hilly terrains, obtained with $10$ different predefined random seeds.

As baseline, we compare the self-attention model (hereon Attention) with:
\begin{enumerate}[label=(\alph*)]
    \item a ``communication-less'' MLP (hereon MLP), that takes the same input as Attention;
    \item a ``communication-based'' MLP (hereon MLP-Comm), that takes as input the local sensor readings and the $4$ values generated by the four adjacent voxels (if any, or zeros, otherwise) at the previous time step.
    Then, it outputs the local actuation and the $4$ values that will be used by the adjacent voxels at the next time step.
    This model is the same as~\cite{medvet2020evolution,medvet2021biodiversity} and, by virtue of the MP mechanism, is an instance of a communication-based controller.
\end{enumerate}

Both models have the same architecture as the self-attention downstream MLP (see \Cref{sec:attention}): as a result, MLP has $41$ parameters for biped and $45$ for comb, while MLP-Comm has $405$ for biped and $440$ for comb.
MLP-Comm and MLP differ from Attention in that they do not employ an attention mechanism to obtain importance scores for the inputs.
For the optimization, we use the same EA of \Cref{sec:ea}.

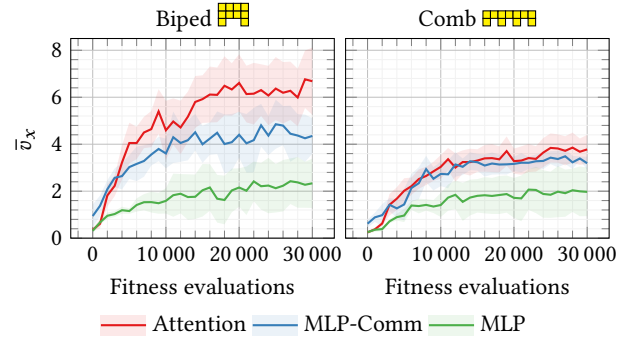
\begin{figure}%[ht!]
    \centering
    \begin{tikzpicture}
        \begin{groupplot}[
            width=0.6\linewidth,
            height=0.5\linewidth,
            group style={
                group size=2 by 1,
                horizontal sep=1.5mm,
                vertical sep=1.5mm,
                xticklabels at=edge bottom,
                yticklabels at=edge left
            },
            every axis plot/.append style={thick},
            scaled x ticks = false,
            grid=both,
            grid style={line width=.1pt, draw=gray!10},
            major grid style={line width=.15pt, draw=gray!50},
            minor tick num=4,
            xlabel={Fitness evaluations},
            ymin=0,ymax=8.5,
            title style={anchor=base, yshift=-0.5mm},
            legend style={draw=none}
        ]
            \nextgroupplot[
                align=center,
                legend columns=3,
                legend entries={Attention,MLP-Comm,MLP},
                legend to name=legendRQ1evo,
                ylabel={$\overline{v}_x$},
                title={Biped \vsr[1mm]{4}{3}{1111-1111-1001}}
            ]
            \linewitherror{data/line/evo.biped.txt}{i}{attn_mu}{attn_std}{cola1}
            \linewitherror{data/line/evo.biped.txt}{i}{neumann_mu}{neumann_std}{cola2}
            \linewitherror{data/line/evo.biped.txt}{i}{base_mu}{base_std}{cola3}
            
            \nextgroupplot[
                title={Comb \vsr[1mm]{7}{2}{1111111-1010101}}
            ]
            \linewitherror{data/line/evo.comb.txt}{i}{attn_mu}{attn_std}{cola1}
            \linewitherror{data/line/evo.comb.txt}{i}{neumann_mu}{neumann_std}{cola2}
            \linewitherror{data/line/evo.comb.txt}{i}{base_mu}{base_std}{cola3}
        \end{groupplot}
    \end{tikzpicture}
    \pgfplotslegendfromname{legendRQ1evo}
    \caption{
        Median $\pm$ standard deviation (solid line and shaded area) of the average velocity for the best individuals found during each evolutionary run, obtained with three controller models and two shapes.
        Attention is never worse than the baselines.
    }
    \label{fig:rq1-evo}
\end{figure}

\begin{figure}
    \centering
    \begin{tikzpicture}
        \begin{groupplot}[
            boxplot,
            boxplot/draw direction=y,
            width=0.6\linewidth,
            height=0.5\linewidth,
            group style={
                group size=2 by 1,
                horizontal sep=1.5mm,
                vertical sep=1.5mm,
                yticklabels at=edge left
            },
            ymin=-1,ymax=13.5,
            legend cell align={left},
            ymajorgrids=true,
            yminorgrids=true,
            grid style={line width=.1pt, draw=gray!10},
            major grid style={line width=.15pt, draw=gray!50},
            minor y tick num=4,
            xmajorticks=false,
            xminorticks=false,
            title style={anchor=base, yshift=-0.5mm},
            legend style={draw=none}
        ]
            \nextgroupplot[
                align=center,
                legend columns=3,
                legend entries={Attention,MLP-Comm,MLP},
                title={Biped \vsr[1mm]{4}{3}{1111-1111-1001}},
                ylabel={$\overline{v}_x$},
                legend to name=legendRQ1val
            ]
            \addlegendimage{area legend,color=cola1,fill}
            \addlegendimage{area legend,color=cola2,fill}
            \addlegendimage{area legend,color=cola3,fill}
            \addplot[black,fill=cola1] table[y=attn] {data/box/val.biped.txt};
            \addplot[black,fill=cola2] table[y=neumann] {data/box/val.biped.txt};
            \addplot[black,fill=cola3] table[y=base] {data/box/val.biped.txt};
            \pvalue{1}{2}{8.5}{$<0.01$}
            \pvalue{2}{3}{8.5}{$<0.01$}
            \pvalue{1}{3}{11.5}{$<0.01$}
            
            \nextgroupplot[
                title={Comb \vsr[1mm]{7}{2}{1111111-1010101}}
            ]
            \addplot[black,fill=cola1] table[y=attn] {data/box/val.comb.txt};
            \addplot[black,fill=cola2] table[y=neumann] {data/box/val.comb.txt};
            \addplot[black,fill=cola3] table[y=base] {data/box/val.comb.txt};
            \pvalue{1}{2}{5.5}{$<0.01$}
            \pvalue{2}{3}{5.5}{$<0.01$}
            \pvalue{1}{3}{8.5}{$<0.01$}
        \end{groupplot}
    \end{tikzpicture}
    \\
    \pgfplotslegendfromname{legendRQ1val}
    \caption{
        Distribution of the average velocity across $10$ unseen hilly terrains for the best individuals found during each evolutionary run, obtained with two controller types and two shapes.
        Attention outperforms the baselines in terms of re-assessment.
    }
    \label{fig:rq1-val}
\end{figure}
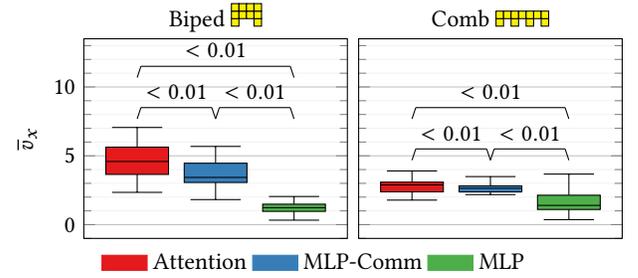

We summarize the results in \Cref{fig:rq1-evo}, which plots $\overline{v}_x$ in terms of median $\pm$ standard deviation for the best individuals over the course of evolution.
Moreover, \Cref{fig:rq1-val} shows $\overline{v}_x$ for the best individuals over the re-assessment terrains.
For the same shape, it also reports the $p$-value for the statistical test against the null hypothesis of equality between the medians.

From the figures, we find that Attention outperforms both baselines.
The curves of \Cref{fig:rq1-evo} also suggest that all models settle on a plateau and that continuing evolution would unlikely bring better results.
Attention individuals also perform better in terms of re-assessment: they achieve significantly better $\overline{v}_x$ on unseen terrains.
Moreover, MLP-Comm outdoes a poor-performing MLP, as attested by the low $\overline{v}_x$ scores, especially for the biped shape.
We visually inspected the evolved behaviors for the best individuals and found them to be highly adapted to a locomotion task on hilly terrain; indeed, bipeds hop on their legs as equines do and combs propagate leg movements from posterior to anterior as millipedes do.
We made videos available at \url{https://softrobots.github.io}\footnote{Number $1$ and number $2$.}.
With no communication, a ``vanilla'' MLP controller fails; it does not enable communication-less VSRs.
To reach a decent performance, we must add a MP communication mechanism.
Intuitively, the MP mechanism is very important for emerging behavior to arise with an MLP-based controller.
Self-attention instead does enable truly communication-less VSRs: it performs better or comparatively to both a communication-less and a communication-based MLP.
We believe the reason for this to be that self-attention is a form of adaptive weights, tailoring importance scores to the inputs and, indirectly, to the particular voxel.

\begin{figure*}
    \centering
    \subcaptionbox{\label{fig:frames-1}}{
        \includegraphics[width=0.18\linewidth]{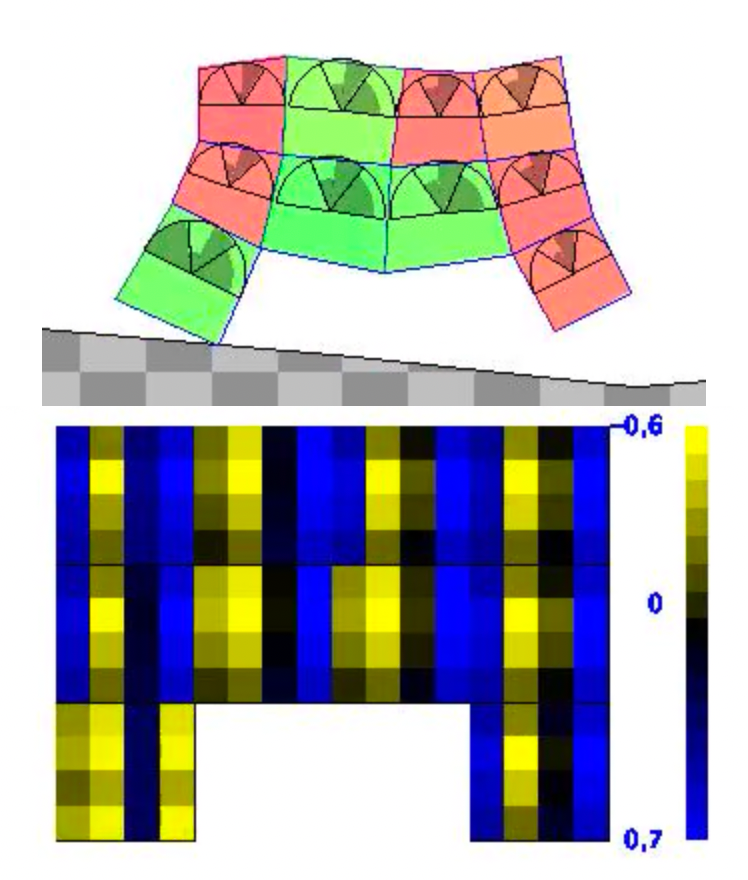}
    }
    \subcaptionbox{\label{fig:frames-2}}{
        \includegraphics[width=0.18\linewidth]{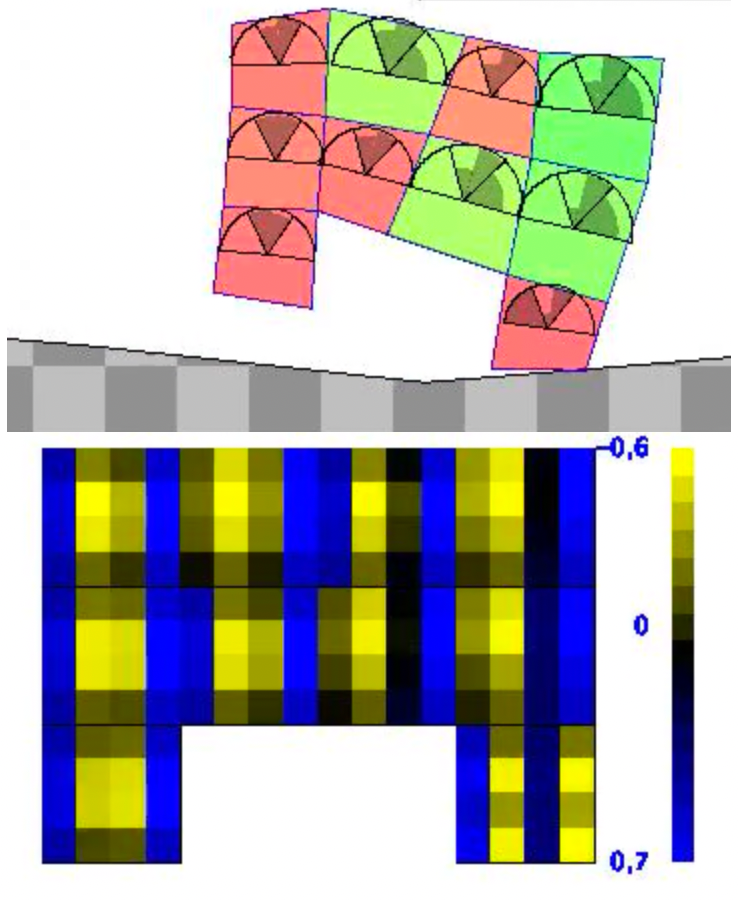}
    }
    \subcaptionbox{\label{fig:frames-3}}{
        \includegraphics[width=0.18\linewidth]{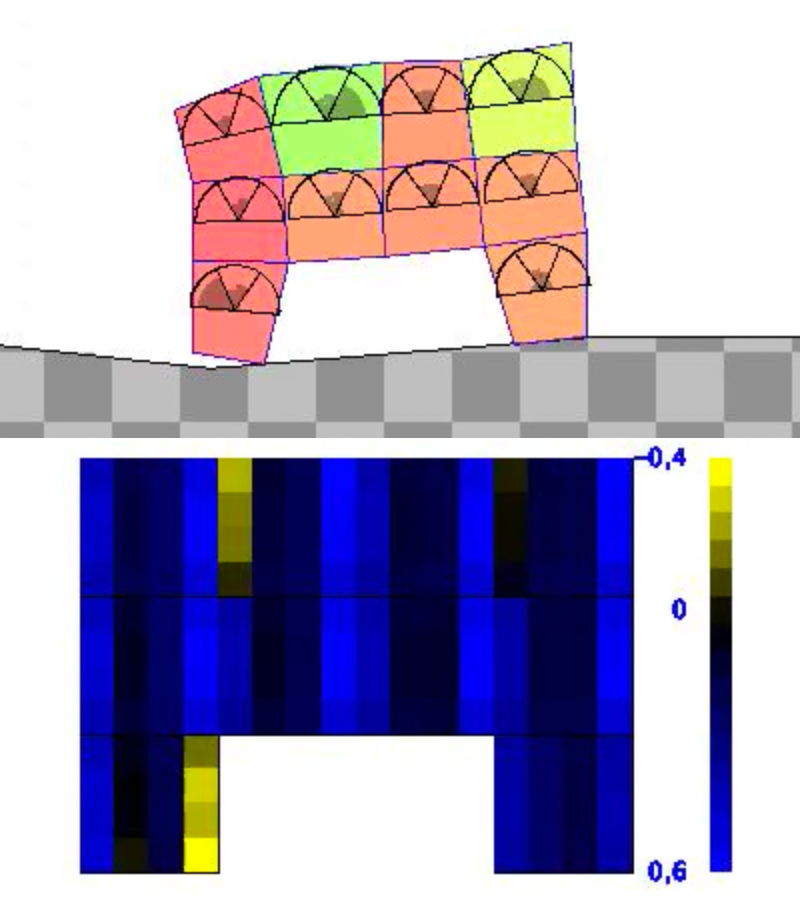}
    }
    \subcaptionbox{\label{fig:frames-4}}{
        \includegraphics[width=0.18\linewidth]{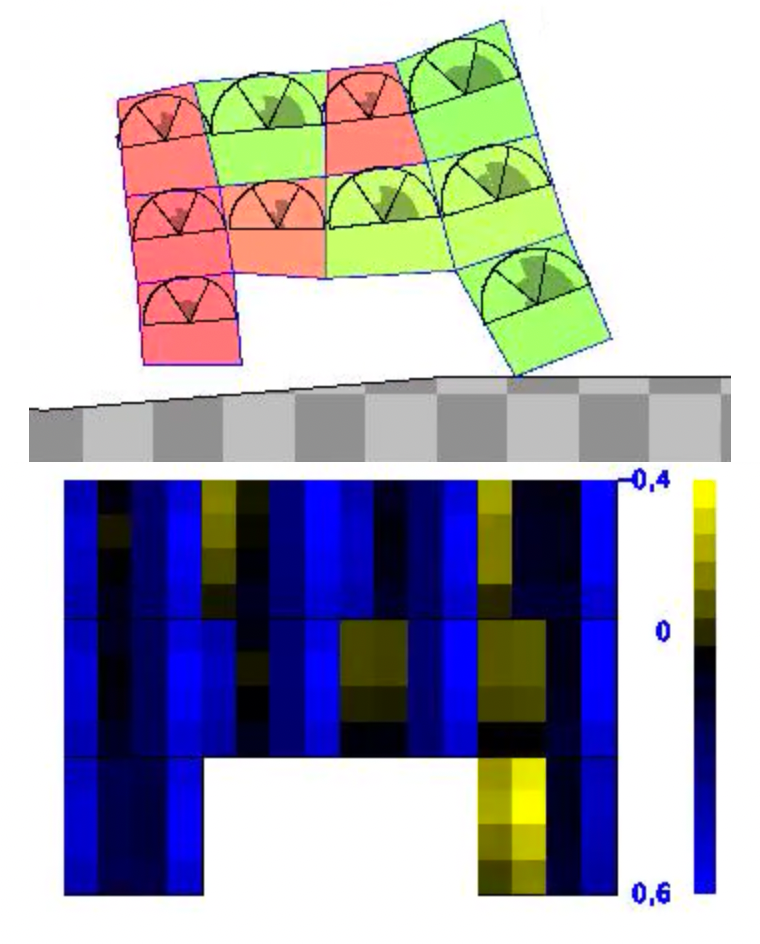}
    }
    \subcaptionbox{\label{fig:frames-5}}{
        \includegraphics[width=0.18\linewidth]{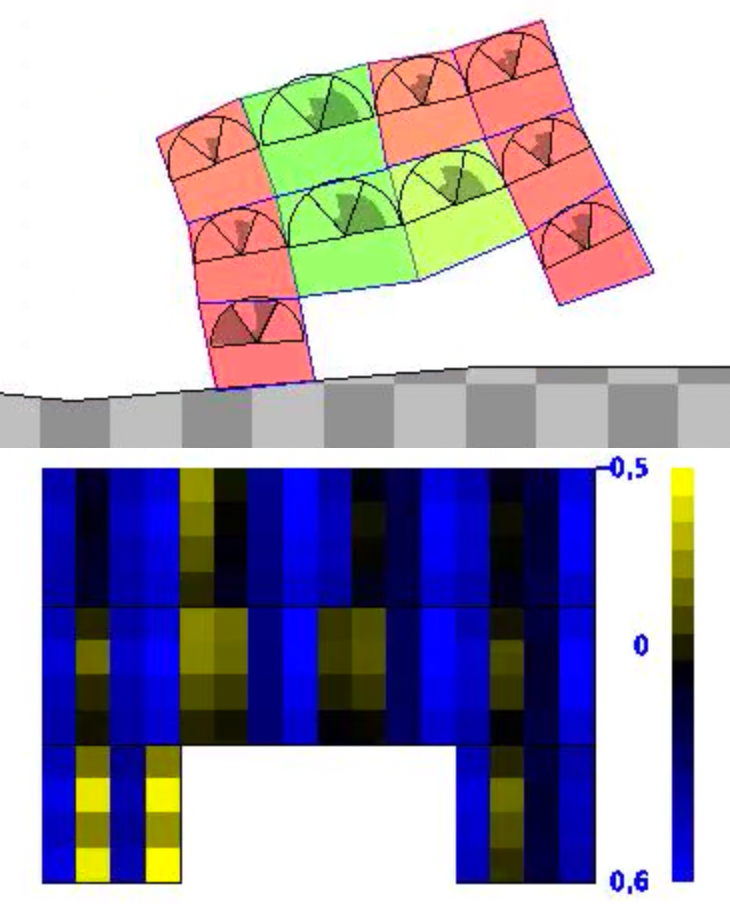}
    }
    \caption{
        Time-lapse showing locomotion for a sample VSR and the corresponding attention matrices.
        The color of each voxel encodes the ratio between its current area and its rest area: red for $<1$, yellow $\approx 1$, green $>1$; the circular sectors drawn at the center of each voxel indicate the current sensor readings.
        The color of each attention score tells the strength of the compatibility between inputs, and provides an interpretation for what is driving the behavior of the VSR.
    }
    \label{fig:rq1-attention-frames}
\end{figure*}

In addition to the evolved behaviors, we also visualized the attention matrices.
Indeed, interpretability is one of the major strengths of self-attention and allows humans to get an insight into the robot inner decision mechanisms.
To illustrate that point, \Cref{fig:rq1-attention-frames} presents a locomotion time-lapse for a sample VSR: at each time step, it displays the robot state in the top row and the corresponding attention matrices in the bottom row; as happens with our distributed controller, there is one such attention matrix for every voxel in the morphology.
Columns and rows of the attention matrices correspond to sensors, in particular: first for the touch sensor, second and third for the $x$- and $y$-velocity sensors, fourth for the area sensor.
The color of each attention score tells the strength of the compatibility between inputs, and provides an interpretation for what is driving the behavior of the VSR.

From the visualization, we realize attention scores are consistent with humans intuition and common sense.
From \Cref{fig:frames-1} to \Cref{fig:frames-2}, attention shifts from the touch sensor of the rear leg to the touch sensor of the front leg.
In \Cref{fig:frames-3}, the VSR stops at a hole in the terrain, and the attention matrix witnesses a radical shift as a consequence.
In \Cref{fig:frames-4} and \Cref{fig:frames-5}, the VSR starts walking again by first focusing on the front leg, and then on the rear leg, initiating locomotion once again.
We found the other individuals to present similar attention patterns.

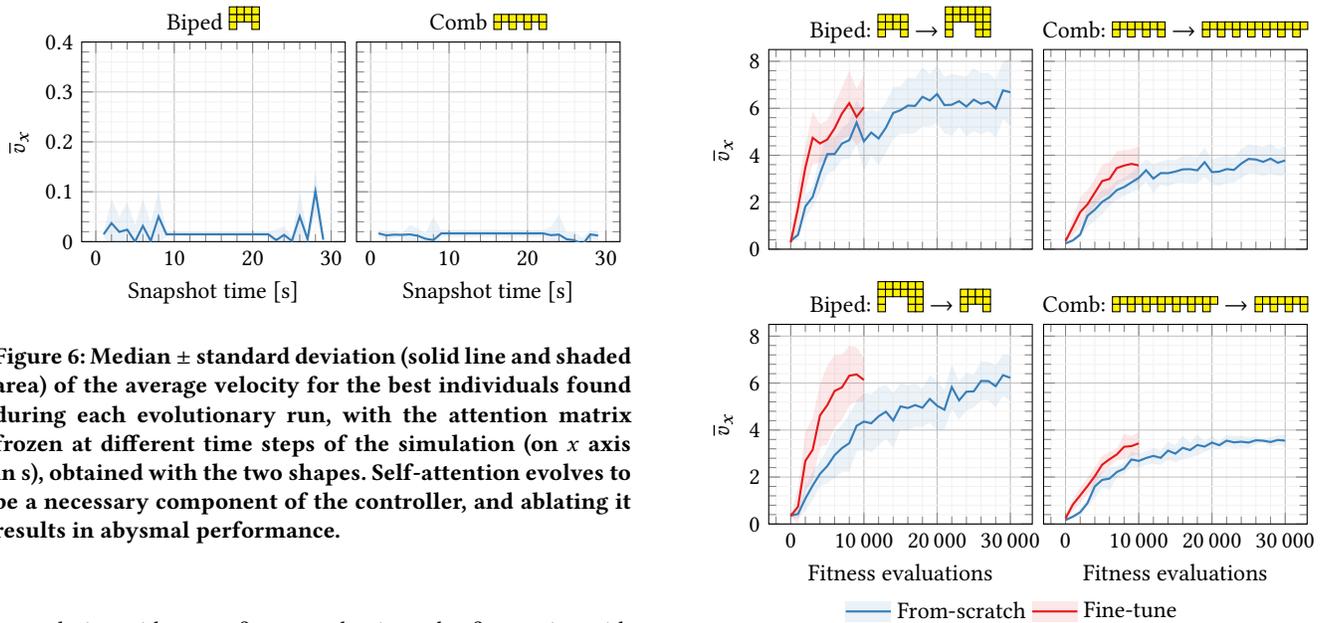
\begin{figure}
    \centering
    \begin{tikzpicture}
        \begin{groupplot}[
            width=0.6\linewidth,
            height=0.5\linewidth,
            grid=both,
            grid style={line width=.1pt, draw=gray!10},
            major grid style={line width=.2pt,draw=gray!50},
            minor tick num=4,
            group style={
                group size=2 by 1,
                horizontal sep=1.5mm,
                vertical sep=1.5mm,
                xticklabels at=edge bottom,
                yticklabels at=edge left
            },
            title style={anchor=base, yshift=-0.5mm},
            ymin=0,ymax=0.4
        ]
            \nextgroupplot[
                title={Biped \vsr[1mm]{4}{3}{1111-1111-1001}},
                ylabel={$\overline{v}_x$},
                xlabel={Snapshot time [\si{\second}]}
            ]
            \linewitherrorsubsam{data/line/abl.biped.txt}{t}{mu}{std}{cola2}
            
            \nextgroupplot[
                title={Comb \vsr[1mm]{7}{2}{1111111-1010101}},
                xlabel={Snapshot time [\si{\second}]}
            ]
            \linewitherrorsubsam{data/line/abl.comb.txt}{t}{mu}{std}{cola2}
        \end{groupplot}
    \end{tikzpicture}
    \caption{
        Median $\pm$ standard deviation (solid line and shaded area) of the average velocity for the best individuals found during each evolutionary run, with the attention matrix frozen at different time steps of the simulation (on $x$ axis in \si{\second}), obtained with the two shapes.
        Self-attention evolves to be a necessary component of the controller, and ablating it results in abysmal performance. 
    }
    \label{fig:rq1-ablation}
\end{figure}

We conducted an ablation study to assess whether evolution found a trivial solution for self-attention or not.
Indeed, AL researchers are all aware of the many uncanny and ``creative'' convergences that artificial evolution is capable of~\cite{lehman2020surprising}.
In particular, we test whether the attention update over time is needed or not, i.e., if there is a ``one attention to rule them all'' case.
To this end, we conducted the following procedure.
Given an evolved VSR with Attention, we take a snapshot of it at every second of the simulation, alongside its attention matrix at that time step.
For every such snapshot, we simulate it on a fixed unseen hilly terrain for $\SI{30}{\second}$ with the attention matrix frozen (i.e., it is the same as the one of the snapshot).
We repeated this procedure for the best individual of every evolutionary run.
We summarize the results in \Cref{fig:rq1-ablation} in terms of $\overline{v}_x$ (median $\pm$ standard deviation), with the time (in \si{\second}) at which we took the corresponding snapshot on the $x$-axis. 
We see that performance drops dramatically, meaning that self-attention is indeed a fundamental piece in the architecture and that it is non-trivial.

Through that evidence, we can answer positively to \ref{item:rq-1}: VSRs evolved with self-attention can solve the task of locomotion and outperform an MLP baseline without relying on inter-module communication, while being fairly able to generalize to unseen terrains.

\subsubsection{\ref{item:rq-2}: why does self-attention work?}
\label{sec:rq-2}
We hypothesize the reasons for self-attention effectiveness to be:
\begin{enumerate}[label=(\alph*)]
    \item self-attention evolves to represent an \emph{inductive bias}, i.e., what tasks the controller is naturally suited at tackling, and
    \item self-attention evolves to be an \emph{instinctive} component (akin to reflexes in biology), that complements decision-making with a fast time-scale of adaptation.
\end{enumerate}

We first focus our attention on the inductive bias hypothesis.
In the field of machine learning, an ``inductive bias'' refers to the set of assumptions made by an algorithm in order to generalize to novel and unseen data~\cite{mitchell1980need}; in our case, data refer to tasks.
If self-attention evolved an inductive bias, as others did point out~\cite{zambaldi2018deep}, it would represent features that are general for the task of locomotion, and are thus suitable for generalization.

To test the inductive bias hypothesis, we conduct an experiment of generalization to unseen morphologies.
Given the attention module $\vec{\theta}_\text{attn}$ evolved on a specific morphology, we wonder whether we can freeze it and use it as an off-the-shelf ``feature extractor'' for a different morphology---as mentioned in \Cref{sec:attention}, $\vec{\theta}_\text{attn}$ dimension is agnostic with respect to the morphology.
If self-attention evolved an inductive bias, it would be useful to control the new morphology.
In particular, for a frozen self-attention, we fine-tune the downstream module.
The goal of fine-tuning is to have a VSR that is effective and converges faster than evolving from scratch.
If that were the case, self-attention would be useful to quickly assemble new VSRs for a different task, starting from pre-optimized components.
This fine-tuning procedure is of crucial importance, since we expect one-shot generalization to fail as a consequence of the embodied cognition paradigm~\cite{pfeifer2006body,shapiro2019embodied}, which posits a deep entanglement between the morphology and the controller of an embodied agent.

For every best individual $\vec{\theta}^\star = \left[\vec{\theta}^\star_\text{attn} \; \vec{\theta}^\star_f \right]$ of an evolutionary run from the experiments of \Cref{sec:rq-1}, we freeze its attention module parameters $\vec{\theta}^\star_\text{attn}$ and fine-tune its downstream module parameters $\vec{\theta}^\star_f$ using the same GA of \Cref{sec:ea} (i.e., mutation and crossover operate on $\vec{\theta}^\star_f$ only), yet starting from a different initial population.
In detail, the initial population is composed of $\vec{\theta}^\star$ and $n_\text{pop}-1$ mutations of it, obtained by copying $\vec{\theta}^\star_\text{attn}$ and re-initializing $\vec{\theta}_f$ by sampling uniformly $[-1,+1]^{|\vec{\theta}_f|}$.

We conducted an experimental campaign of $5$ evolutionary runs, lasting $n_\text{evals}=\num{20000}$ fitness evaluations each, with one larger version of each of the two shapes used in the previous experiments.
We fine-tuned the best individual of every run as explained above, freezing and transferring its attention module to the corresponding smaller or larger morphology, i.e., from biped-small \vsr[1mm]{4}{3}{1111-1111-1001} to biped-large \vsr[1mm]{6}{4}{111111-111111-100011-100011}, from biped-large to biped-small, from comb-small \vsr[1mm]{7}{2}{1111111-1010101} to comb-large \vsr[1mm]{14}{2}{11111111111111-10101010101010}, and from comb-large to comb-small.
For fine-tuning, we set $n_\text{evals}=\num{10000}$ fitness evaluations as termination criterion, as our goal is to quickly adapt to a new task starting from pre-optimized components (i.e., the attention module).
We summarize the results in \Cref{fig:rq2-inductive}: for each morphology, we compare $\overline{v}_x$ (in terms of median $\pm$ standard deviation) for fine-tuning (Fine-tune) and from-scratch optimization (From-scratch).
For a fair comparison, we plot the results for From-scratch over \num{30000} fitness evaluations, since every Fine-tune run is the outcome of an evolution with \num{20000} fitness evaluations plus fine-tuning with \num{10000}.

\begin{figure}
    \centering
    \begin{tikzpicture}
        \begin{groupplot}[
            width=0.6\linewidth,
            height=0.5\linewidth,
            group style={
                group size=2 by 2,
                horizontal sep=1.5mm,
                vertical sep=10mm,
                xticklabels at=edge bottom,
                yticklabels at=edge left
            },
            every axis plot/.append style={thick},
            scaled x ticks = false,
            grid=both,
            grid style={line width=.1pt, draw=gray!10},
            major grid style={line width=.15pt, draw=gray!50},
            minor tick num=4,
            title style={anchor=base, yshift=-0.5mm},
            ymin=0,ymax=8.5,
            legend style={draw=none}
        ]
            \nextgroupplot[
                title={Biped: \vsr[1mm]{4}{3}{1111-1111-1001} $\rightarrow$ \vsr[1mm]{6}{4}{111111-111111-100011-100011}},
                ylabel={$\overline{v}_x$},
                align=center,
                legend columns=2,
                legend entries={From-scratch,Fine-tune},
                legend to name=legendRQ2inductive
            ]
            \linewitherror{data/line/best.biped-4x3.txt}{i}{mu}{std}{cola2}
            \linewitherror{data/line/finetune.biped-4x3.txt}{i}{mu}{std}{cola1}

            \nextgroupplot[
                title={Comb: \vsr[1mm]{7}{2}{1111111-1010101} $\rightarrow$ \vsr[1mm]{14}{2}{11111111111111-10101010101010}}
            ]
            \linewitherror{data/line/best.comb-7x2.txt}{i}{mu}{std}{cola2}
            \linewitherror{data/line/finetune.comb-7x2.txt}{i}{mu}{std}{cola1}
            
            \nextgroupplot[
                title={Biped: \vsr[1mm]{6}{4}{111111-111111-100011-100011} $\rightarrow$ \vsr[1mm]{4}{3}{1111-1111-1001}},
                xlabel={Fitness evaluations},
                ylabel={$\overline{v}_x$},
            ]
            \linewitherror{data/line/best.biped-6x4.txt}{i}{mu}{std}{cola2}
            \linewitherror{data/line/finetune.biped-6x4.txt}{i}{mu}{std}{cola1}
            
            \nextgroupplot[
                title={Comb: \vsr[1mm]{14}{2}{11111111111111-10101010101010} $\rightarrow$ \vsr[1mm]{7}{2}{1111111-1010101}},
                xlabel={Fitness evaluations}
            ]
            \linewitherror{data/line/best.comb-14x2.txt}{i}{mu}{std}{cola2}
            \linewitherror{data/line/finetune.comb-14x2.txt}{i}{mu}{std}{cola1}
        \end{groupplot}
    \end{tikzpicture}
    \pgfplotslegendfromname{legendRQ2inductive}
    \caption{
        Median $\pm$ standard deviation (line and shaded area) of the average velocity for the best individuals found during each evolutionary run, with two controller types (color), with optimization from scratch (dashed line) or fine-tuning a pre-optimized attention (solid line) on a different morphology.
        Attention evolves an inductive bias that, generally, represents invariant features for the locomotion task.
    }
    \label{fig:rq2-inductive}
\end{figure}

From \Cref{fig:rq2-inductive}, we see that Fine-tune succeeds in converging much faster.
Moreover, the curves suggest that Fine-tune would benefit from a longer re-optimization; however, we remark that the goal of this study is not to reach the best performance, but to converge cheaply.
We visually inspected the behaviors of the Fine-tune individuals and found them to be adapted for a locomotion task and consistent with the behaviors observed in \Cref{sec:rq-1}.
We made videos available at \url{https://softrobots.github.io} for the best Fine-tune individuals\footnote{Numbers $3$, $4$, $5$, and $6$.}.

We conclude that it is often possible to re-use an evolved attention module for a smaller or larger morphology, after a fine-tuning stage.
That result is hopeful in the context of future autonomous robotic ecosystems: we expect new robots to be assembled from pre-optimized components, so that they are, at the same time, proficient at new tasks and computationally cheaper than optimizing from scratch.

We believe the result is relevant, as, to date,~\cite{kriegman2021scale} is the only other work addressing generalization to different soft robot morphologies by means of EC.
Albeit ground-breaking, \citet{kriegman2021scale} achieved generalization by computing the fitness function over three different morphologies at every evaluation, which might not be feasible in a fully autonomous setting.
We achieve generalization with no penalty during evolution, at the cost of introducing a fine-tuning stage.
Finally, we remark it could be argued that other works, e.g.,~\cite{wang2018nervenet,pathak2019learning,huang2020one}, achieved generalization to unseen morphologies via techniques more sample-efficient than EC, most notably Reinforcement Learning (RL).
While those advances are indeed noteworthy, there are reasons to use EC in the first place: empirically, we find that RL algorithms are notoriously more unstable than evolution; indeed, there is a recent body of literature that shows how even simple EAs can achieve performance competitive to state-of-the-art RL systems~\cite{salimans2017evolution,such2017deep,risi2019deep}, at the benefit of less complexity and less sensitivity to hyperparameters.

It is worth noting that evolution and learning can complement each other very effectively~\cite{eiben2020if}.
For example, in evolutionary robotics, learning augments evolution by allowing newborn controllers to adapt more quickly to their bodies~\cite{gupta2021embodied,luo2021effects}.
It is clear that self-attention is not a form of learning: in particular, there is no memory or state, as every attention matrix is computed just from current observations.
We argue that self-attention evolves to have an instantaneous time-scale of adaptation; similarly to reflexes, it immediately reacts to sensory perceptions.
Under this light, self-attention belongs to the ``System 1'' mode of thought: as put forward in the seminal work of \citet{kahnemen2011thinking}, System 1 thinking is fast, instinctive, and emotional.
System 1 then exists at the level of the sub-conscious.
That hypothesis is in line with the nature of attention in biology~\cite{cohen1991attention,treisman1992automaticity}.

To validate that hypothesis, we perform an experiment by slowing down the dynamics of attention.
In particular, given the attention matrix $\vec{A}$ as defined in \Cref{eq:self-attention}, that depends only on current input, we define the matrix $\vec{A}^{\prime(k)}$ as:
\begin{equation}
    \label{eq:slow-attention}
    \vec{A}^{\prime(k)} =
    \begin{cases}
        \alpha\vec{A} + \eta\vec{A}^{\prime(k-1)} &\text{if}\ k > 0\\
        \vec{A} &\text{if}\ k = 0
    \end{cases}
\end{equation}
where $\alpha, \eta \in [0,1]$.
$\alpha$ acts as a \emph{learning rate}, weighting the current attention, while $\eta$ acts as a \emph{forget rate}, weighting the attention compounded from the previous time steps.
In the following, we substitute $\vec{A}$ with $\vec{A}^{\prime(k)}$ in \Cref{eq:attention}, i.e., $\vec{A}^{\prime(k)}$ is the effective attention matrix at time step $k$, i.e., at time $t = k\Delta t$.
As a side note, if we enforce $\alpha + \eta = 1$, \Cref{eq:slow-attention} resembles the ``associative weights'' memory of \citet{ba2016using}.

In the following, we evolve $\alpha$ and $\eta$ as part of the genotype.
To enforce $\alpha, \eta \in [0,1]$, when mapping a genotype $\vec{\theta}$ into a robot phenotype, we set both to be the absolute value clipped at $1$ of the corresponding genes.
In doing so, we let evolution discover what is the fittest value for $\alpha$ and $\eta$ and, as a result, what is the optimal balance between current information and past information in self-attention.
In a certain sense, we are ``meta-evolving'' self-attention.

\begin{figure}
    \centering
    \begin{tikzpicture}
        \begin{groupplot}[
            width=0.6\linewidth,
            height=0.5\linewidth,
            group style={
                group size=2 by 1,
                horizontal sep=1.5mm,
                vertical sep=1.5mm,
                xticklabels at=edge bottom,
                yticklabels at=edge left
            },
            every axis plot/.append style={thick},
            scaled x ticks = false,
            grid=both,
            grid style={line width=.1pt, draw=gray!10},
            major grid style={line width=.15pt, draw=gray!50},
            minor tick num=4,
            xlabel={Fitness evaluations},
            ymin=-0.25,ymax=1.25,
            title style={anchor=base, yshift=-0.5mm},
            legend style={draw=none}
        ]
            \nextgroupplot[
                title={Biped \vsr[1mm]{4}{3}{1111-1111-1001}},
                ylabel={$\alpha,\eta$},
                align=center,
                legend columns=2,
                legend entries={$\alpha$,$\eta$},
                legend to name=legendRQ2instinct
            ]
            \linewitherror{data/line/ae.biped.txt}{i}{a_mu}{a_std}{cola1}
            \linewitherror{data/line/ae.biped.txt}{i}{e_mu}{e_std}{cola2}
            
            \nextgroupplot[
                title={Comb \vsr[1mm]{7}{2}{1111111-1010101}}
            ]
            \linewitherror{data/line/ae.comb.txt}{i}{a_mu}{a_std}{cola1}
            \linewitherror{data/line/ae.comb.txt}{i}{e_mu}{e_std}{cola2}
        \end{groupplot}
    \end{tikzpicture}
    \pgfplotslegendfromname{legendRQ2instinct}
    \caption{
        Median $\pm$ standard deviation (solid line and shaded area) of the $\alpha$ and $\eta$ values of \Cref{eq:slow-attention} for the best individuals found during each evolutionary run, obtained with two shapes.
        Self-attention evolves to quickly forget past information and attend to the present.
    }
    \label{fig:rq2-compound}
\end{figure}
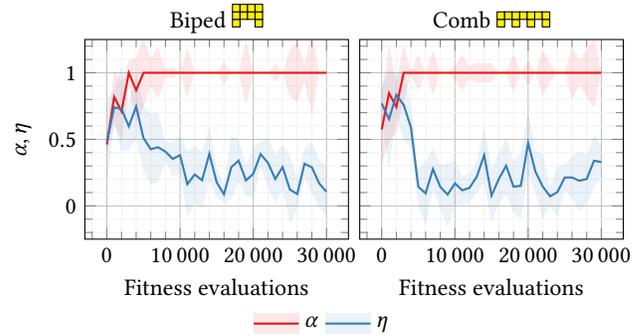

With those settings, we conducted an experimental campaign of $5$ evolutionary runs with the shapes of \Cref{sec:rq-1}.
We found the resulting best individuals to be not significantly different from those of \Cref{sec:rq-1}, in terms of effectiveness and qualitative analysis of self-attention, so we simply show in \Cref{fig:rq2-compound} the evolution of $\alpha$ and $\eta$.
We notice a clear trend: $\alpha$ evolves to be $1$, while $\eta$ evolves to approach $0$.
In other words, self-attention evolves to keep all the present information and retain little from the past.
Those results mostly confirm our hypothesis that evolution leads self-attention to have an instantaneous time-scale of adaptation, akin to an instinct.

Through those tests, we can answer to \ref{item:rq-2}: self-attention works because it evolves to represent an inductive bias that is useful for generalization; at the same time, self-attention evolves to be an instinctive adaptation mechanism, allowing the controller to react quickly to changes in the inputs.

\section{Conclusion}
\label{sec:conclusion}
In the path toward autonomous robotic ecosystems, full exploitation of modularity remains a roadblock~\cite{yim2007modular}.
Minimizing the need for communication among robot modules would greatly facilitate the disassembly of robots in components and their reassembly in different forms, to cope, e.g., with different tasks.

Considering the case of VSRs, that, a priori, enjoy a high degree of modularity (in both bodies and brains), those points is precisely what the paper demonstrates:
\begin{enumerate}[label=(\roman*)]
    \item we can evolve controllers that dispense with inter-module communication for a locomotion task on hilly terrains;
    \item we can achieve generalization to unseen morphologies, after a short fine-tuning with evolution.
\end{enumerate}
To do so, we employ a local self-attention mechanism---a form of adaptive weights---to let every voxel adapt to its local perceptions, and optimize the controller parameters (which are the same for all the voxels) with EC.
On the other side, an MLP baseline is not proficient in the same task.
We also observe that self-attention evolves to be an instinctive adaptation mechanism, allowing the controller to quickly react to input changes.

Yet, it is unclear how our results extend beyond the scope of modular soft robots, including modular rigid robots.
Future work will develop on these ideas to investigate how critical the soft body dynamics are in supporting full controller modularity, and how to leverage self-attention to evolve communication patterns tailored to different voxels.

\begin{acks}
The experiments in this work were conducted using virtual machines provided by Google Cloud Platform granted to F.P.
\end{acks}

\bibliographystyle{ACM-Reference-Format}
\bibliography{bibliography}

\end{document}

%% file: plot-macros.tex
\usepackage{tikz,pgfplots,pgfplotstable}
\usetikzlibrary{patterns}
\pgfplotsset{compat=newest}
\usepgfplotslibrary{groupplots,fillbetween,statistics}
\newcommand{\linewitherror}[5]{
    \addplot [each nth point=1,name path=minuserror,draw=none,no markers,forget plot] table [x={#2},y expr=\thisrow{#3}-\thisrow{#4}] {#1};
    \addplot [each nth point=1,name path=pluserror,draw=none,no markers,forget plot] table [x={#2},y expr=\thisrow{#3}+\thisrow{#4}] {#1};
    \addplot [forget plot,fill=#5,opacity=0.1] fill between[on layer={},of=pluserror and minuserror];
    \addplot [each nth point=1,#5,thick,no markers,legend image code/.code={\fill [fill=#5, opacity=0.1, draw=none] (0mm,-1ex) -- (0mm,1ex) -- (6mm,1ex) -- (6mm,-1ex) -- cycle; \draw [#5,thick] (0mm,0mm) -- (6mm,0mm);}] table [x={#2},y={#3}] {#1};
}

\newcommand{\linewitherrorsubsam}[5]{
    \addplot [each nth point=1,name path=minuserror,draw=none,no markers,forget plot] table [x={#2},y expr=\thisrow{#3}-\thisrow{#4}] {#1};
    \addplot [each nth point=1,name path=pluserror,draw=none,no markers,forget plot] table [x={#2},y expr=\thisrow{#3}+\thisrow{#4}] {#1};
    \addplot [forget plot,fill=#5,opacity=0.1] fill between[on layer={},of=pluserror and minuserror];
    \addplot [each nth point=1,#5,thick,no markers,legend image code/.code={\fill [fill=#5, opacity=0.1, draw=none] (0mm,-1ex) -- (0mm,1ex) -- (6mm,1ex) -- (6mm,-1ex) -- cycle; \draw [#5,thick] (0mm,0mm) -- (6mm,0mm);}] table [x={#2},y={#3}] {#1};
}

\newcommand{\pvalue}[4]{
    \draw ([xshift=.5mm]#1,#3) -- ([yshift=-1.5mm]#1,#3);
    \draw ([xshift=-.5mm]#2,#3) -- ([yshift=-1.5mm]#2,#3);
    \draw ([xshift=.5mm]#1,#3) -- node [above,scale=1] {#4} ([xshift=-.5mm]#2,#3);
}

\def\addlegendimage{\csname pgfplots@addlegendimage\endcsname}

\definecolor{cola1}{RGB}{228,26,28}
\definecolor{cola2}{RGB}{55,126,184}
\definecolor{cola3}{RGB}{77,175,74}
\definecolor{cola4}{RGB}{152,78,163}
\definecolor{cola5}{RGB}{255,127,0}
\definecolor{cola6}{RGB}{255,255,51}
\definecolor{cola7}{RGB}{166,86,40}

\definecolor{colb1}{RGB}{102,194,165}
\definecolor{colb2}{RGB}{252,141,98}
\definecolor{colb3}{RGB}{141,160,203}
\definecolor{colb4}{RGB}{231,138,195}

%% file: main.bbl

\begin{thebibliography}{79}

%%% the Web address.

\ifx \showCODEN    \undefined \def \showCODEN     #1{\unskip}     \fi
\ifx \showDOI      \undefined \def \showDOI       #1{#1}\fi
\ifx \showISBNx    \undefined \def \showISBNx     #1{\unskip}     \fi
\ifx \showISBNxiii \undefined \def \showISBNxiii  #1{\unskip}     \fi
\ifx \showISSN     \undefined \def \showISSN      #1{\unskip}     \fi
\ifx \showLCCN     \undefined \def \showLCCN      #1{\unskip}     \fi
\ifx \shownote     \undefined \def \shownote      #1{#1}          \fi
\ifx \showarticletitle \undefined \def \showarticletitle #1{#1}   \fi
\ifx \showURL      \undefined \def \showURL       {\relax}        \fi
% invisible to TeX
\providecommand\bibfield[2]{#2}
\providecommand\bibinfo[2]{#2}
\providecommand\natexlab[1]{#1}
\providecommand\showeprint[2][]{arXiv:#2}

\bibitem[\protect\citeauthoryear{Ba, Hinton, Mnih, Leibo, and Ionescu}{Ba
  et~al\mbox{.}}{2016}]%
        {ba2016using}
\bibfield{author}{\bibinfo{person}{Jimmy Ba}, \bibinfo{person}{Geoffrey~E
  Hinton}, \bibinfo{person}{Volodymyr Mnih}, \bibinfo{person}{Joel~Z Leibo},
  {and} \bibinfo{person}{Catalin Ionescu}.} \bibinfo{year}{2016}\natexlab{}.
\newblock \showarticletitle{Using fast weights to attend to the recent past}.
\newblock \bibinfo{journal}{\emph{Advances in Neural Information Processing
  Systems}}  \bibinfo{volume}{29} (\bibinfo{year}{2016}),
  \bibinfo{pages}{4331--4339}.
\newblock


\bibitem[\protect\citeauthoryear{Bahdanau, Cho, and Bengio}{Bahdanau
  et~al\mbox{.}}{2014}]%
        {bahdanau2014neural}
\bibfield{author}{\bibinfo{person}{Dzmitry Bahdanau},
  \bibinfo{person}{Kyunghyun Cho}, {and} \bibinfo{person}{Yoshua Bengio}.}
  \bibinfo{year}{2014}\natexlab{}.
\newblock \showarticletitle{Neural machine translation by jointly learning to
  align and translate}.
\newblock \bibinfo{journal}{\emph{arXiv preprint arXiv:1409.0473}}
  (\bibinfo{year}{2014}).
\newblock


\bibitem[\protect\citeauthoryear{Bhatia, Jackson, Tian, Xu, and Matusik}{Bhatia
  et~al\mbox{.}}{2021}]%
        {bhatia2021evolution}
\bibfield{author}{\bibinfo{person}{Jagdeep Bhatia}, \bibinfo{person}{Holly
  Jackson}, \bibinfo{person}{Yunsheng Tian}, \bibinfo{person}{Jie Xu}, {and}
  \bibinfo{person}{Wojciech Matusik}.} \bibinfo{year}{2021}\natexlab{}.
\newblock \showarticletitle{Evolution gym: A large-scale benchmark for evolving
  soft robots}.
\newblock \bibinfo{journal}{\emph{Advances in Neural Information Processing
  Systems}}  \bibinfo{volume}{34} (\bibinfo{year}{2021}).
\newblock


\bibitem[\protect\citeauthoryear{Blackiston, Lederer, Kriegman, Garnier,
  Bongard, and Levin}{Blackiston et~al\mbox{.}}{2021}]%
        {blackiston2021cellular}
\bibfield{author}{\bibinfo{person}{Douglas Blackiston}, \bibinfo{person}{Emma
  Lederer}, \bibinfo{person}{Sam Kriegman}, \bibinfo{person}{Simon Garnier},
  \bibinfo{person}{Joshua Bongard}, {and} \bibinfo{person}{Michael Levin}.}
  \bibinfo{year}{2021}\natexlab{}.
\newblock \showarticletitle{A cellular platform for the development of
  synthetic living machines}.
\newblock \bibinfo{journal}{\emph{Science Robotics}} \bibinfo{volume}{6},
  \bibinfo{number}{52} (\bibinfo{year}{2021}), \bibinfo{pages}{eabf1571}.
\newblock


\bibitem[\protect\citeauthoryear{Buchanan, Le~Goff, Li, Hart, Eiben, De~Carlo,
  Winfield, Hale, Woolley, Angus, et~al\mbox{.}}{Buchanan
  et~al\mbox{.}}{2020}]%
        {buchanan2020bootstrapping}
\bibfield{author}{\bibinfo{person}{Edgar Buchanan}, \bibinfo{person}{L{\'e}ni~K
  Le~Goff}, \bibinfo{person}{Wei Li}, \bibinfo{person}{Emma Hart},
  \bibinfo{person}{Agoston~E Eiben}, \bibinfo{person}{Matteo De~Carlo},
  \bibinfo{person}{Alan~F Winfield}, \bibinfo{person}{Matthew~F Hale},
  \bibinfo{person}{Robert Woolley}, \bibinfo{person}{Mike Angus},
  {et~al\mbox{.}}} \bibinfo{year}{2020}\natexlab{}.
\newblock \showarticletitle{Bootstrapping Artificial Evolution to Design Robots
  for Autonomous Fabrication}.
\newblock \bibinfo{journal}{\emph{Robotics}} \bibinfo{volume}{9},
  \bibinfo{number}{4} (\bibinfo{year}{2020}), \bibinfo{pages}{106}.
\newblock


\bibitem[\protect\citeauthoryear{Chaudhari, Mithal, Polatkan, and
  Ramanath}{Chaudhari et~al\mbox{.}}{2021}]%
        {chaudhari2021attentive}
\bibfield{author}{\bibinfo{person}{Sneha Chaudhari}, \bibinfo{person}{Varun
  Mithal}, \bibinfo{person}{Gungor Polatkan}, {and} \bibinfo{person}{Rohan
  Ramanath}.} \bibinfo{year}{2021}\natexlab{}.
\newblock \showarticletitle{An attentive survey of attention models}.
\newblock \bibinfo{journal}{\emph{ACM Transactions on Intelligent Systems and
  Technology (TIST)}} \bibinfo{volume}{12}, \bibinfo{number}{5}
  (\bibinfo{year}{2021}), \bibinfo{pages}{1--32}.
\newblock


\bibitem[\protect\citeauthoryear{Cheney, MacCurdy, Clune, and Lipson}{Cheney
  et~al\mbox{.}}{2014}]%
        {cheney2014unshackling}
\bibfield{author}{\bibinfo{person}{Nick Cheney}, \bibinfo{person}{Robert
  MacCurdy}, \bibinfo{person}{Jeff Clune}, {and} \bibinfo{person}{Hod Lipson}.}
  \bibinfo{year}{2014}\natexlab{}.
\newblock \showarticletitle{Unshackling evolution: evolving soft robots with
  multiple materials and a powerful generative encoding}.
\newblock \bibinfo{journal}{\emph{ACM SIGEVOlution}} \bibinfo{volume}{7},
  \bibinfo{number}{1} (\bibinfo{year}{2014}), \bibinfo{pages}{11--23}.
\newblock


\bibitem[\protect\citeauthoryear{Choi, Lee, and Zhang}{Choi
  et~al\mbox{.}}{2017}]%
        {choi2017multi}
\bibfield{author}{\bibinfo{person}{Jinyoung Choi}, \bibinfo{person}{Beom-Jin
  Lee}, {and} \bibinfo{person}{Byoung-Tak Zhang}.}
  \bibinfo{year}{2017}\natexlab{}.
\newblock \showarticletitle{Multi-focus attention network for efficient deep
  reinforcement learning}. In \bibinfo{booktitle}{\emph{Workshops at the
  thirty-first AAAI conference on artificial intelligence}}.
\newblock


\bibitem[\protect\citeauthoryear{Christensen, Schultz, and Stoy}{Christensen
  et~al\mbox{.}}{2013}]%
        {christensen2013distributed}
\bibfield{author}{\bibinfo{person}{David~Johan Christensen},
  \bibinfo{person}{Ulrik~Pagh Schultz}, {and} \bibinfo{person}{Kasper Stoy}.}
  \bibinfo{year}{2013}\natexlab{}.
\newblock \showarticletitle{A distributed and morphology-independent strategy
  for adaptive locomotion in self-reconfigurable modular robots}.
\newblock \bibinfo{journal}{\emph{Robotics and Autonomous Systems}}
  \bibinfo{volume}{61}, \bibinfo{number}{9} (\bibinfo{year}{2013}),
  \bibinfo{pages}{1021--1035}.
\newblock


\bibitem[\protect\citeauthoryear{Cohen and Rafal}{Cohen and Rafal}{1991}]%
        {cohen1991attention}
\bibfield{author}{\bibinfo{person}{Asher Cohen} {and} \bibinfo{person}{Robert~D
  Rafal}.} \bibinfo{year}{1991}\natexlab{}.
\newblock \showarticletitle{Attention and feature integration: Illusory
  conjunctions in a patient with a parietal lobe lesion}.
\newblock \bibinfo{journal}{\emph{Psychological science}} \bibinfo{volume}{2},
  \bibinfo{number}{2} (\bibinfo{year}{1991}), \bibinfo{pages}{106--110}.
\newblock


\bibitem[\protect\citeauthoryear{Cussat-Blanc, Harrington, and
  Banzhaf}{Cussat-Blanc et~al\mbox{.}}{2019}]%
        {cussat2019artificial}
\bibfield{author}{\bibinfo{person}{Sylvain Cussat-Blanc}, \bibinfo{person}{Kyle
  Harrington}, {and} \bibinfo{person}{Wolfgang Banzhaf}.}
  \bibinfo{year}{2019}\natexlab{}.
\newblock \showarticletitle{Artificial gene regulatory networks—a review}.
\newblock \bibinfo{journal}{\emph{Artificial life}} \bibinfo{volume}{24},
  \bibinfo{number}{4} (\bibinfo{year}{2019}), \bibinfo{pages}{296--328}.
\newblock


\bibitem[\protect\citeauthoryear{De~Jong}{De~Jong}{2006}]%
        {de2006evolutionary}
\bibfield{author}{\bibinfo{person}{Kenneth~A De~Jong}.}
  \bibinfo{year}{2006}\natexlab{}.
\newblock \bibinfo{booktitle}{\emph{{Evolutionary Computation: A Unified
  Approach}}}.
\newblock \bibinfo{publisher}{MIT Press}.
\newblock


\bibitem[\protect\citeauthoryear{Devlin, Chang, Lee, and Toutanova}{Devlin
  et~al\mbox{.}}{2018}]%
        {devlin2018bert}
\bibfield{author}{\bibinfo{person}{Jacob Devlin}, \bibinfo{person}{Ming-Wei
  Chang}, \bibinfo{person}{Kenton Lee}, {and} \bibinfo{person}{Kristina
  Toutanova}.} \bibinfo{year}{2018}\natexlab{}.
\newblock \showarticletitle{Bert: Pre-training of deep bidirectional
  transformers for language understanding}.
\newblock \bibinfo{journal}{\emph{arXiv preprint arXiv:1810.04805}}
  (\bibinfo{year}{2018}).
\newblock


\bibitem[\protect\citeauthoryear{Dosovitskiy, Beyer, Kolesnikov, Weissenborn,
  Zhai, Unterthiner, Dehghani, Minderer, Heigold, Gelly,
  et~al\mbox{.}}{Dosovitskiy et~al\mbox{.}}{2020}]%
        {dosovitskiy2020image}
\bibfield{author}{\bibinfo{person}{Alexey Dosovitskiy}, \bibinfo{person}{Lucas
  Beyer}, \bibinfo{person}{Alexander Kolesnikov}, \bibinfo{person}{Dirk
  Weissenborn}, \bibinfo{person}{Xiaohua Zhai}, \bibinfo{person}{Thomas
  Unterthiner}, \bibinfo{person}{Mostafa Dehghani}, \bibinfo{person}{Matthias
  Minderer}, \bibinfo{person}{Georg Heigold}, \bibinfo{person}{Sylvain Gelly},
  {et~al\mbox{.}}} \bibinfo{year}{2020}\natexlab{}.
\newblock \showarticletitle{An image is worth 16x16 words: Transformers for
  image recognition at scale}.
\newblock \bibinfo{journal}{\emph{arXiv preprint arXiv:2010.11929}}
  (\bibinfo{year}{2020}).
\newblock


\bibitem[\protect\citeauthoryear{Driver}{Driver}{2001}]%
        {driver2001selective}
\bibfield{author}{\bibinfo{person}{Jon Driver}.}
  \bibinfo{year}{2001}\natexlab{}.
\newblock \showarticletitle{A selective review of selective attention research
  from the past century}.
\newblock \bibinfo{journal}{\emph{British Journal of Psychology}}
  \bibinfo{volume}{92}, \bibinfo{number}{1} (\bibinfo{year}{2001}),
  \bibinfo{pages}{53--78}.
\newblock


\bibitem[\protect\citeauthoryear{Eiben}{Eiben}{2021}]%
        {eiben2021real}
\bibfield{author}{\bibinfo{person}{AE Eiben}.} \bibinfo{year}{2021}\natexlab{}.
\newblock \showarticletitle{Real-World Robot Evolution: Why Would it (not)
  Work?}
\newblock \bibinfo{journal}{\emph{Frontiers in Robotics and AI}}
  (\bibinfo{year}{2021}), \bibinfo{pages}{243}.
\newblock


\bibitem[\protect\citeauthoryear{Eiben and Hart}{Eiben and Hart}{2020}]%
        {eiben2020if}
\bibfield{author}{\bibinfo{person}{AE Eiben} {and} \bibinfo{person}{Emma
  Hart}.} \bibinfo{year}{2020}\natexlab{}.
\newblock \showarticletitle{If it evolves it needs to learn}. In
  \bibinfo{booktitle}{\emph{Proceedings of the 2020 Genetic and Evolutionary
  Computation Conference Companion}}. \bibinfo{pages}{1383--1384}.
\newblock


\bibitem[\protect\citeauthoryear{Fai{\~n}a}{Fai{\~n}a}{2021}]%
        {faina2021evolving}
\bibfield{author}{\bibinfo{person}{Andres Fai{\~n}a}.}
  \bibinfo{year}{2021}\natexlab{}.
\newblock \showarticletitle{Evolving Modular Robots: Challenges and
  Opportunities}. In \bibinfo{booktitle}{\emph{ALIFE 2021: The 2021 Conference
  on Artificial Life}}. MIT Press.
\newblock


\bibitem[\protect\citeauthoryear{Ferigo, Iacca, Medvet, and Pigozzi}{Ferigo
  et~al\mbox{.}}{2021}]%
        {ferigo2021evolving}
\bibfield{author}{\bibinfo{person}{Andrea Ferigo}, \bibinfo{person}{Giovanni
  Iacca}, \bibinfo{person}{Eric Medvet}, {and} \bibinfo{person}{Federico
  Pigozzi}.} \bibinfo{year}{2021}\natexlab{}.
\newblock \showarticletitle{Evolving Hebbian Learning Rules in Voxel-based Soft
  Robots}.
\newblock  (\bibinfo{year}{2021}).
\newblock


\bibitem[\protect\citeauthoryear{Fodor}{Fodor}{1983}]%
        {fodor1983modularity}
\bibfield{author}{\bibinfo{person}{Jerry~A Fodor}.}
  \bibinfo{year}{1983}\natexlab{}.
\newblock \bibinfo{booktitle}{\emph{The modularity of mind}}.
\newblock \bibinfo{publisher}{MIT press}.
\newblock


\bibitem[\protect\citeauthoryear{Galassi, Lippi, and Torroni}{Galassi
  et~al\mbox{.}}{2020}]%
        {galassi2020attention}
\bibfield{author}{\bibinfo{person}{Andrea Galassi}, \bibinfo{person}{Marco
  Lippi}, {and} \bibinfo{person}{Paolo Torroni}.}
  \bibinfo{year}{2020}\natexlab{}.
\newblock \showarticletitle{Attention in natural language processing}.
\newblock \bibinfo{journal}{\emph{IEEE Transactions on Neural Networks and
  Learning Systems}} (\bibinfo{year}{2020}).
\newblock


\bibitem[\protect\citeauthoryear{Grattarola, Livi, and Alippi}{Grattarola
  et~al\mbox{.}}{2021}]%
        {grattarola2021learning}
\bibfield{author}{\bibinfo{person}{Daniele Grattarola},
  \bibinfo{person}{Lorenzo Livi}, {and} \bibinfo{person}{Cesare Alippi}.}
  \bibinfo{year}{2021}\natexlab{}.
\newblock \showarticletitle{Learning graph cellular automata}.
\newblock \bibinfo{journal}{\emph{Advances in Neural Information Processing
  Systems}}  \bibinfo{volume}{34} (\bibinfo{year}{2021}).
\newblock


\bibitem[\protect\citeauthoryear{Gro{\ss}, Tuci, Dorigo, Bonani, and
  Mondada}{Gro{\ss} et~al\mbox{.}}{2006}]%
        {gross2006object}
\bibfield{author}{\bibinfo{person}{Roderich Gro{\ss}}, \bibinfo{person}{Elio
  Tuci}, \bibinfo{person}{Marco Dorigo}, \bibinfo{person}{Michael Bonani},
  {and} \bibinfo{person}{Francesco Mondada}.} \bibinfo{year}{2006}\natexlab{}.
\newblock \showarticletitle{Object transport by modular robots that
  self-assemble}. In \bibinfo{booktitle}{\emph{Proceedings 2006 IEEE
  International Conference on Robotics and Automation, 2006. ICRA 2006.}} IEEE,
  \bibinfo{pages}{2558--2564}.
\newblock


\bibitem[\protect\citeauthoryear{Gupta, Savarese, Ganguli, and Fei-Fei}{Gupta
  et~al\mbox{.}}{2021}]%
        {gupta2021embodied}
\bibfield{author}{\bibinfo{person}{Agrim Gupta}, \bibinfo{person}{Silvio
  Savarese}, \bibinfo{person}{Surya Ganguli}, {and} \bibinfo{person}{Li
  Fei-Fei}.} \bibinfo{year}{2021}\natexlab{}.
\newblock \showarticletitle{{Embodied Intelligence via Learning and
  Evolution}}.
\newblock \bibinfo{journal}{\emph{arXiv preprint arXiv:2102.02202}}
  (\bibinfo{year}{2021}).
\newblock


\bibitem[\protect\citeauthoryear{Ha}{Ha}{2019}]%
        {ha2019reinforcement}
\bibfield{author}{\bibinfo{person}{David Ha}.} \bibinfo{year}{2019}\natexlab{}.
\newblock \showarticletitle{Reinforcement learning for improving agent design}.
\newblock \bibinfo{journal}{\emph{Artificial life}} \bibinfo{volume}{25},
  \bibinfo{number}{4} (\bibinfo{year}{2019}), \bibinfo{pages}{352--365}.
\newblock


\bibitem[\protect\citeauthoryear{Ha and Tang}{Ha and Tang}{2021}]%
        {ha2021collective}
\bibfield{author}{\bibinfo{person}{David Ha} {and} \bibinfo{person}{Yujin
  Tang}.} \bibinfo{year}{2021}\natexlab{}.
\newblock \showarticletitle{Collective Intelligence for Deep Learning: A Survey
  of Recent Developments}.
\newblock \bibinfo{journal}{\emph{arXiv preprint arXiv:2111.14377}}
  (\bibinfo{year}{2021}).
\newblock


\bibitem[\protect\citeauthoryear{Hale, Angus, Buchanan, Li, Woolley, Le~Goff,
  De~Carlo, Timmis, Winfield, Hart, et~al\mbox{.}}{Hale et~al\mbox{.}}{2020}]%
        {hale2020hardware}
\bibfield{author}{\bibinfo{person}{Matthew~F Hale}, \bibinfo{person}{Mike
  Angus}, \bibinfo{person}{Edgar Buchanan}, \bibinfo{person}{Wei Li},
  \bibinfo{person}{Robert Woolley}, \bibinfo{person}{L{\'e}ni~K Le~Goff},
  \bibinfo{person}{Matteo De~Carlo}, \bibinfo{person}{Jon Timmis},
  \bibinfo{person}{Alan~F Winfield}, \bibinfo{person}{Emma Hart},
  {et~al\mbox{.}}} \bibinfo{year}{2020}\natexlab{}.
\newblock \showarticletitle{Hardware design for autonomous robot evolution}. In
  \bibinfo{booktitle}{\emph{2020 IEEE Symposium Series on Computational
  Intelligence (SSCI)}}. IEEE, \bibinfo{pages}{2140--2147}.
\newblock


\bibitem[\protect\citeauthoryear{Hauser, Ijspeert, F{\"u}chslin, Pfeifer, and
  Maass}{Hauser et~al\mbox{.}}{2011}]%
        {hauser2011towards}
\bibfield{author}{\bibinfo{person}{Helmut Hauser}, \bibinfo{person}{Auke~J
  Ijspeert}, \bibinfo{person}{Rudolf~M F{\"u}chslin}, \bibinfo{person}{Rolf
  Pfeifer}, {and} \bibinfo{person}{Wolfgang Maass}.}
  \bibinfo{year}{2011}\natexlab{}.
\newblock \showarticletitle{Towards a theoretical foundation for morphological
  computation with compliant bodies}.
\newblock \bibinfo{journal}{\emph{Biological cybernetics}}
  \bibinfo{volume}{105}, \bibinfo{number}{5} (\bibinfo{year}{2011}),
  \bibinfo{pages}{355--370}.
\newblock


\bibitem[\protect\citeauthoryear{Hiller and Lipson}{Hiller and Lipson}{2012}]%
        {hiller2012automatic}
\bibfield{author}{\bibinfo{person}{Jonathan Hiller} {and} \bibinfo{person}{Hod
  Lipson}.} \bibinfo{year}{2012}\natexlab{}.
\newblock \showarticletitle{Automatic design and manufacture of soft robots}.
\newblock \bibinfo{journal}{\emph{IEEE Transactions on Robotics}}
  \bibinfo{volume}{28}, \bibinfo{number}{2} (\bibinfo{year}{2012}),
  \bibinfo{pages}{457--466}.
\newblock


\bibitem[\protect\citeauthoryear{Horibe, Walker, and Risi}{Horibe
  et~al\mbox{.}}{2021}]%
        {horibe2021regenerating}
\bibfield{author}{\bibinfo{person}{Kazuya Horibe}, \bibinfo{person}{Kathryn
  Walker}, {and} \bibinfo{person}{Sebastian Risi}.}
  \bibinfo{year}{2021}\natexlab{}.
\newblock \showarticletitle{Regenerating Soft Robots Through Neural Cellular
  Automata.}. In \bibinfo{booktitle}{\emph{EuroGP}}. \bibinfo{pages}{36--50}.
\newblock


\bibitem[\protect\citeauthoryear{Howison, Hauser, Hughes, and Iida}{Howison
  et~al\mbox{.}}{2020}]%
        {howison2020reality}
\bibfield{author}{\bibinfo{person}{Toby Howison}, \bibinfo{person}{Simon
  Hauser}, \bibinfo{person}{Josie Hughes}, {and} \bibinfo{person}{Fumiya
  Iida}.} \bibinfo{year}{2020}\natexlab{}.
\newblock \showarticletitle{Reality-assisted evolution of soft robots through
  large-scale physical experimentation: a review}.
\newblock \bibinfo{journal}{\emph{Artificial Life}} \bibinfo{volume}{26},
  \bibinfo{number}{4} (\bibinfo{year}{2020}), \bibinfo{pages}{484--506}.
\newblock


\bibitem[\protect\citeauthoryear{Huang, Mordatch, and Pathak}{Huang
  et~al\mbox{.}}{2020}]%
        {huang2020one}
\bibfield{author}{\bibinfo{person}{Wenlong Huang}, \bibinfo{person}{Igor
  Mordatch}, {and} \bibinfo{person}{Deepak Pathak}.}
  \bibinfo{year}{2020}\natexlab{}.
\newblock \showarticletitle{One policy to control them all: Shared modular
  policies for agent-agnostic control}. In
  \bibinfo{booktitle}{\emph{International Conference on Machine Learning}}.
  PMLR, \bibinfo{pages}{4455--4464}.
\newblock


\bibitem[\protect\citeauthoryear{Joachimczak, Suzuki, and Arita}{Joachimczak
  et~al\mbox{.}}{2016}]%
        {joachimczak2016metamrphosis}
\bibfield{author}{\bibinfo{person}{Michał Joachimczak}, \bibinfo{person}{Reiji
  Suzuki}, {and} \bibinfo{person}{Takaya Arita}.}
  \bibinfo{year}{2016}\natexlab{}.
\newblock \showarticletitle{Artificial Metamorphosis: Evolutionary Design of
  Transforming, Soft-Bodied Robots}.
\newblock \bibinfo{journal}{\emph{Artificial Life}} \bibinfo{volume}{22},
  \bibinfo{number}{3} (\bibinfo{year}{2016}), \bibinfo{pages}{271--298}.
\newblock
\urldef\tempurl%
\url{https://doi.org/10.1162/ARTL_a_00207}
\showURL{%
\tempurl}
\newblock
\shownote{PMID: 27139940.}


\bibitem[\protect\citeauthoryear{Kahneman}{Kahneman}{2011}]%
        {kahnemen2011thinking}
\bibfield{author}{\bibinfo{person}{Daniel Kahneman}.}
  \bibinfo{year}{2011}\natexlab{}.
\newblock \showarticletitle{Thinking fast and slow}.
\newblock \bibinfo{journal}{\emph{New York: Farrar, Straus and Giroux}}
  (\bibinfo{year}{2011}).
\newblock


\bibitem[\protect\citeauthoryear{Kalat, Faal, and Onal}{Kalat
  et~al\mbox{.}}{2018}]%
        {kalat2018decentralized}
\bibfield{author}{\bibinfo{person}{Shadi~Tasdighi Kalat},
  \bibinfo{person}{Siamak~G Faal}, {and} \bibinfo{person}{Cagdas~D Onal}.}
  \bibinfo{year}{2018}\natexlab{}.
\newblock \showarticletitle{A decentralized, communication-free force
  distribution method with application to collective object manipulation}.
\newblock \bibinfo{journal}{\emph{Journal of Dynamic Systems, Measurement, and
  Control}} \bibinfo{volume}{140}, \bibinfo{number}{9} (\bibinfo{year}{2018}),
  \bibinfo{pages}{091012}.
\newblock


\bibitem[\protect\citeauthoryear{Kamimura, Kurokawa, Toshida, Tomita, Murata,
  and Kokaji}{Kamimura et~al\mbox{.}}{2003}]%
        {kamimura2003automatic}
\bibfield{author}{\bibinfo{person}{Akiya Kamimura}, \bibinfo{person}{Haruhisa
  Kurokawa}, \bibinfo{person}{E Toshida}, \bibinfo{person}{Kohji Tomita},
  \bibinfo{person}{Satoshi Murata}, {and} \bibinfo{person}{Shigeru Kokaji}.}
  \bibinfo{year}{2003}\natexlab{}.
\newblock \showarticletitle{Automatic locomotion pattern generation for modular
  robots}. In \bibinfo{booktitle}{\emph{2003 IEEE International Conference on
  Robotics and Automation (Cat. No. 03CH37422)}}, Vol.~\bibinfo{volume}{1}.
  IEEE, \bibinfo{pages}{714--720}.
\newblock


\bibitem[\protect\citeauthoryear{Khan, Naseer, Hayat, Zamir, Khan, and
  Shah}{Khan et~al\mbox{.}}{2021}]%
        {khan2021transformers}
\bibfield{author}{\bibinfo{person}{Salman Khan}, \bibinfo{person}{Muzammal
  Naseer}, \bibinfo{person}{Munawar Hayat}, \bibinfo{person}{Syed~Waqas Zamir},
  \bibinfo{person}{Fahad~Shahbaz Khan}, {and} \bibinfo{person}{Mubarak Shah}.}
  \bibinfo{year}{2021}\natexlab{}.
\newblock \showarticletitle{Transformers in vision: A survey}.
\newblock \bibinfo{journal}{\emph{arXiv preprint arXiv:2101.01169}}
  (\bibinfo{year}{2021}).
\newblock


\bibitem[\protect\citeauthoryear{Kriegman, Blackiston, Levin, and
  Bongard}{Kriegman et~al\mbox{.}}{2020}]%
        {kriegman2020scalable}
\bibfield{author}{\bibinfo{person}{Sam Kriegman}, \bibinfo{person}{Douglas
  Blackiston}, \bibinfo{person}{Michael Levin}, {and} \bibinfo{person}{Josh
  Bongard}.} \bibinfo{year}{2020}\natexlab{}.
\newblock \showarticletitle{A scalable pipeline for designing reconfigurable
  organisms}.
\newblock \bibinfo{journal}{\emph{Proceedings of the National Academy of
  Sciences}} \bibinfo{volume}{117}, \bibinfo{number}{4} (\bibinfo{year}{2020}),
  \bibinfo{pages}{1853--1859}.
\newblock


\bibitem[\protect\citeauthoryear{Kriegman, Cheney, and Bongard}{Kriegman
  et~al\mbox{.}}{2018}]%
        {kriegman2018morphological}
\bibfield{author}{\bibinfo{person}{Sam Kriegman}, \bibinfo{person}{Nick
  Cheney}, {and} \bibinfo{person}{Josh Bongard}.}
  \bibinfo{year}{2018}\natexlab{}.
\newblock \showarticletitle{How morphological development can guide evolution}.
\newblock \bibinfo{journal}{\emph{Scientific reports}} \bibinfo{volume}{8},
  \bibinfo{number}{1} (\bibinfo{year}{2018}), \bibinfo{pages}{13934}.
\newblock


\bibitem[\protect\citeauthoryear{Kriegman, Nasab, Blackiston, Steele, Levin,
  Kramer-Bottiglio, and Bongard}{Kriegman et~al\mbox{.}}{2021}]%
        {kriegman2021scale}
\bibfield{author}{\bibinfo{person}{Sam Kriegman},
  \bibinfo{person}{Amir~Mohammadi Nasab}, \bibinfo{person}{Douglas Blackiston},
  \bibinfo{person}{Hannah Steele}, \bibinfo{person}{Michael Levin},
  \bibinfo{person}{Rebecca Kramer-Bottiglio}, {and} \bibinfo{person}{Josh
  Bongard}.} \bibinfo{year}{2021}\natexlab{}.
\newblock \showarticletitle{Scale invariant robot behavior with fractals}.
\newblock \bibinfo{journal}{\emph{arXiv preprint arXiv:2103.04876}}
  (\bibinfo{year}{2021}).
\newblock


\bibitem[\protect\citeauthoryear{Lan, De~Carlo, van Diggelen, Tomczak, Roijers,
  and Eiben}{Lan et~al\mbox{.}}{2021}]%
        {lan2021learning}
\bibfield{author}{\bibinfo{person}{Gongjin Lan}, \bibinfo{person}{Matteo
  De~Carlo}, \bibinfo{person}{Fuda van Diggelen}, \bibinfo{person}{Jakub~M
  Tomczak}, \bibinfo{person}{Diederik~M Roijers}, {and}
  \bibinfo{person}{Agoston~E Eiben}.} \bibinfo{year}{2021}\natexlab{}.
\newblock \showarticletitle{Learning directed locomotion in modular robots with
  evolvable morphologies}.
\newblock \bibinfo{journal}{\emph{Applied Soft Computing}}
  \bibinfo{volume}{111} (\bibinfo{year}{2021}), \bibinfo{pages}{107688}.
\newblock


\bibitem[\protect\citeauthoryear{Lee, Lee, Kim, Kosiorek, Choi, and Teh}{Lee
  et~al\mbox{.}}{2019}]%
        {lee2019set}
\bibfield{author}{\bibinfo{person}{Juho Lee}, \bibinfo{person}{Yoonho Lee},
  \bibinfo{person}{Jungtaek Kim}, \bibinfo{person}{Adam Kosiorek},
  \bibinfo{person}{Seungjin Choi}, {and} \bibinfo{person}{Yee~Whye Teh}.}
  \bibinfo{year}{2019}\natexlab{}.
\newblock \showarticletitle{Set transformer: A framework for attention-based
  permutation-invariant neural networks}. In
  \bibinfo{booktitle}{\emph{International Conference on Machine Learning}}.
  PMLR, \bibinfo{pages}{3744--3753}.
\newblock


\bibitem[\protect\citeauthoryear{Lehman, Clune, Misevic, Adami, Altenberg,
  Beaulieu, Bentley, Bernard, Beslon, Bryson, et~al\mbox{.}}{Lehman
  et~al\mbox{.}}{2020}]%
        {lehman2020surprising}
\bibfield{author}{\bibinfo{person}{Joel Lehman}, \bibinfo{person}{Jeff Clune},
  \bibinfo{person}{Dusan Misevic}, \bibinfo{person}{Christoph Adami},
  \bibinfo{person}{Lee Altenberg}, \bibinfo{person}{Julie Beaulieu},
  \bibinfo{person}{Peter~J Bentley}, \bibinfo{person}{Samuel Bernard},
  \bibinfo{person}{Guillaume Beslon}, \bibinfo{person}{David~M Bryson},
  {et~al\mbox{.}}} \bibinfo{year}{2020}\natexlab{}.
\newblock \showarticletitle{The surprising creativity of digital evolution: A
  collection of anecdotes from the evolutionary computation and artificial life
  research communities}.
\newblock \bibinfo{journal}{\emph{Artificial life}} \bibinfo{volume}{26},
  \bibinfo{number}{2} (\bibinfo{year}{2020}), \bibinfo{pages}{274--306}.
\newblock


\bibitem[\protect\citeauthoryear{Luo, Stuurman, Tomczak, Ellers, and Eiben}{Luo
  et~al\mbox{.}}{2021}]%
        {luo2021effects}
\bibfield{author}{\bibinfo{person}{Jie Luo}, \bibinfo{person}{Aart Stuurman},
  \bibinfo{person}{Jakub~M Tomczak}, \bibinfo{person}{Jacintha Ellers}, {and}
  \bibinfo{person}{Agoston~E Eiben}.} \bibinfo{year}{2021}\natexlab{}.
\newblock \showarticletitle{The Effects of Learning in Morphologically Evolving
  Robot Systems}.
\newblock \bibinfo{journal}{\emph{arXiv preprint arXiv:2111.09851}}
  (\bibinfo{year}{2021}).
\newblock


\bibitem[\protect\citeauthoryear{Luong, Pham, and Manning}{Luong
  et~al\mbox{.}}{2015}]%
        {luong2015effective}
\bibfield{author}{\bibinfo{person}{Minh-Thang Luong}, \bibinfo{person}{Hieu
  Pham}, {and} \bibinfo{person}{Christopher~D Manning}.}
  \bibinfo{year}{2015}\natexlab{}.
\newblock \showarticletitle{Effective approaches to attention-based neural
  machine translation}.
\newblock \bibinfo{journal}{\emph{arXiv preprint arXiv:1508.04025}}
  (\bibinfo{year}{2015}).
\newblock


\bibitem[\protect\citeauthoryear{Martius, Der, and Ay}{Martius
  et~al\mbox{.}}{2013}]%
        {martius2013information}
\bibfield{author}{\bibinfo{person}{Georg Martius}, \bibinfo{person}{Ralf Der},
  {and} \bibinfo{person}{Nihat Ay}.} \bibinfo{year}{2013}\natexlab{}.
\newblock \showarticletitle{Information driven self-organization of complex
  robotic behaviors}.
\newblock \bibinfo{journal}{\emph{PloS one}} \bibinfo{volume}{8},
  \bibinfo{number}{5} (\bibinfo{year}{2013}), \bibinfo{pages}{e63400}.
\newblock


\bibitem[\protect\citeauthoryear{Medvet, Bartoli, De~Lorenzo, and Fidel}{Medvet
  et~al\mbox{.}}{2020a}]%
        {medvet2020evolution}
\bibfield{author}{\bibinfo{person}{Eric Medvet}, \bibinfo{person}{Alberto
  Bartoli}, \bibinfo{person}{Andrea De~Lorenzo}, {and} \bibinfo{person}{Giulio
  Fidel}.} \bibinfo{year}{2020}\natexlab{a}.
\newblock \showarticletitle{Evolution of distributed neural controllers for
  voxel-based soft robots}. In \bibinfo{booktitle}{\emph{Proceedings of the
  2020 Genetic and Evolutionary Computation Conference}}.
  \bibinfo{pages}{112--120}.
\newblock


\bibitem[\protect\citeauthoryear{Medvet, Bartoli, De~Lorenzo, and
  Seriani}{Medvet et~al\mbox{.}}{2020b}]%
        {medvet20202d}
\bibfield{author}{\bibinfo{person}{Eric Medvet}, \bibinfo{person}{Alberto
  Bartoli}, \bibinfo{person}{Andrea De~Lorenzo}, {and} \bibinfo{person}{Stefano
  Seriani}.} \bibinfo{year}{2020}\natexlab{b}.
\newblock \showarticletitle{{2D-VSR-Sim: A simulation tool for the optimization
  of 2-D voxel-based soft robots}}.
\newblock \bibinfo{journal}{\emph{{SoftwareX}}}  \bibinfo{volume}{12}
  (\bibinfo{year}{2020}).
\newblock


\bibitem[\protect\citeauthoryear{Medvet, Bartoli, Pigozzi, and Rochelli}{Medvet
  et~al\mbox{.}}{2021}]%
        {medvet2021biodiversity}
\bibfield{author}{\bibinfo{person}{Eric Medvet}, \bibinfo{person}{Alberto
  Bartoli}, \bibinfo{person}{Federico Pigozzi}, {and} \bibinfo{person}{Marco
  Rochelli}.} \bibinfo{year}{2021}\natexlab{}.
\newblock \showarticletitle{{Biodiversity in evolved voxel-based soft robots}}.
  In \bibinfo{booktitle}{\emph{{Proceedings of the Genetic and Evolutionary
  Computation Conference}}}. \bibinfo{pages}{129--137}.
\newblock


\bibitem[\protect\citeauthoryear{Mitchell}{Mitchell}{1980}]%
        {mitchell1980need}
\bibfield{author}{\bibinfo{person}{Tom~M Mitchell}.}
  \bibinfo{year}{1980}\natexlab{}.
\newblock \bibinfo{booktitle}{\emph{The need for biases in learning
  generalizations}}.
\newblock \bibinfo{publisher}{Department of Computer Science, Laboratory for
  Computer Science Research~…}.
\newblock


\bibitem[\protect\citeauthoryear{Mordvintsev, Randazzo, Niklasson, and
  Levin}{Mordvintsev et~al\mbox{.}}{2020}]%
        {mordvintsev2020growing}
\bibfield{author}{\bibinfo{person}{Alexander Mordvintsev},
  \bibinfo{person}{Ettore Randazzo}, \bibinfo{person}{Eyvind Niklasson}, {and}
  \bibinfo{person}{Michael Levin}.} \bibinfo{year}{2020}\natexlab{}.
\newblock \showarticletitle{Growing Neural Cellular Automata}.
\newblock \bibinfo{journal}{\emph{Distill}} (\bibinfo{year}{2020}).
\newblock
\urldef\tempurl%
\url{https://doi.org/10.23915/distill.00023}
\showDOI{\tempurl}
\newblock
\shownote{https://distill.pub/2020/growing-ca.}


\bibitem[\protect\citeauthoryear{Nolfi and Floreano}{Nolfi and
  Floreano}{2000}]%
        {nolfi2000evolutionary}
\bibfield{author}{\bibinfo{person}{Stefano Nolfi} {and} \bibinfo{person}{Dario
  Floreano}.} \bibinfo{year}{2000}\natexlab{}.
\newblock \bibinfo{booktitle}{\emph{Evolutionary robotics: The biology,
  intelligence, and technology of self-organizing machines}}.
\newblock \bibinfo{publisher}{MIT press}.
\newblock


\bibitem[\protect\citeauthoryear{Owaki, Horikiri, Nishii, and Ishiguro}{Owaki
  et~al\mbox{.}}{2021}]%
        {owaki2021tegotae}
\bibfield{author}{\bibinfo{person}{Dai Owaki}, \bibinfo{person}{Shun-ya
  Horikiri}, \bibinfo{person}{Jun Nishii}, {and} \bibinfo{person}{Akio
  Ishiguro}.} \bibinfo{year}{2021}\natexlab{}.
\newblock \showarticletitle{Tegotae-Based Control Produces Adaptive Inter-and
  Intra-limb Coordination in Bipedal Walking}.
\newblock \bibinfo{journal}{\emph{Frontiers in neurorobotics}}
  \bibinfo{volume}{15} (\bibinfo{year}{2021}), \bibinfo{pages}{47}.
\newblock


\bibitem[\protect\citeauthoryear{Pathak, Lu, Darrell, Isola, and Efros}{Pathak
  et~al\mbox{.}}{2019}]%
        {pathak2019learning}
\bibfield{author}{\bibinfo{person}{Deepak Pathak}, \bibinfo{person}{Chris Lu},
  \bibinfo{person}{Trevor Darrell}, \bibinfo{person}{Phillip Isola}, {and}
  \bibinfo{person}{Alexei~A Efros}.} \bibinfo{year}{2019}\natexlab{}.
\newblock \showarticletitle{Learning to control self-assembling morphologies: a
  study of generalization via modularity}.
\newblock \bibinfo{journal}{\emph{arXiv preprint arXiv:1902.05546}}
  (\bibinfo{year}{2019}).
\newblock


\bibitem[\protect\citeauthoryear{Pfeifer and Bongard}{Pfeifer and
  Bongard}{2006}]%
        {pfeifer2006body}
\bibfield{author}{\bibinfo{person}{Rolf Pfeifer} {and} \bibinfo{person}{Josh
  Bongard}.} \bibinfo{year}{2006}\natexlab{}.
\newblock \bibinfo{booktitle}{\emph{How the body shapes the way we think: a new
  view of intelligence}}.
\newblock \bibinfo{publisher}{MIT press}.
\newblock


\bibitem[\protect\citeauthoryear{Pigozzi and Medvet}{Pigozzi and
  Medvet}{2022}]%
        {pigozzi2022embodied}
\bibfield{author}{\bibinfo{person}{Federico Pigozzi} {and}
  \bibinfo{person}{Eric Medvet}.} \bibinfo{year}{2022}\natexlab{}.
\newblock \showarticletitle{Evolving Modularity in Soft Robots through an
  Embodied and Self-Organizing Neural Controller}.
\newblock \bibinfo{journal}{\emph{Artificial Life}} (\bibinfo{year}{2022}).
\newblock


\bibitem[\protect\citeauthoryear{Queralta, McCord, Gia, Tenhunen, and
  Westerlund}{Queralta et~al\mbox{.}}{2019}]%
        {queralta2019communication}
\bibfield{author}{\bibinfo{person}{J~Pena Queralta}, \bibinfo{person}{Cassandra
  McCord}, \bibinfo{person}{Tuan~Nguyen Gia}, \bibinfo{person}{Hannu Tenhunen},
  {and} \bibinfo{person}{Tomi Westerlund}.} \bibinfo{year}{2019}\natexlab{}.
\newblock \showarticletitle{Communication-free and index-free distributed
  formation control algorithm for multi-robot systems}.
\newblock \bibinfo{journal}{\emph{Procedia Computer Science}}
  \bibinfo{volume}{151} (\bibinfo{year}{2019}), \bibinfo{pages}{431--438}.
\newblock


\bibitem[\protect\citeauthoryear{Risi and Stanley}{Risi and Stanley}{2019}]%
        {risi2019deep}
\bibfield{author}{\bibinfo{person}{Sebastian Risi} {and}
  \bibinfo{person}{Kenneth~O Stanley}.} \bibinfo{year}{2019}\natexlab{}.
\newblock \showarticletitle{Deep neuroevolution of recurrent and discrete world
  models}. In \bibinfo{booktitle}{\emph{Proceedings of the Genetic and
  Evolutionary Computation Conference}}. \bibinfo{pages}{456--462}.
\newblock


\bibitem[\protect\citeauthoryear{Rus and Tolley}{Rus and Tolley}{2015}]%
        {rus2015design}
\bibfield{author}{\bibinfo{person}{Daniela Rus} {and}
  \bibinfo{person}{Michael~T Tolley}.} \bibinfo{year}{2015}\natexlab{}.
\newblock \showarticletitle{Design, fabrication and control of soft robots}.
\newblock \bibinfo{journal}{\emph{Nature}} \bibinfo{volume}{521},
  \bibinfo{number}{7553} (\bibinfo{year}{2015}), \bibinfo{pages}{467}.
\newblock


\bibitem[\protect\citeauthoryear{Salimans, Ho, Chen, Sidor, and
  Sutskever}{Salimans et~al\mbox{.}}{2017}]%
        {salimans2017evolution}
\bibfield{author}{\bibinfo{person}{Tim Salimans}, \bibinfo{person}{Jonathan
  Ho}, \bibinfo{person}{Xi Chen}, \bibinfo{person}{Szymon Sidor}, {and}
  \bibinfo{person}{Ilya Sutskever}.} \bibinfo{year}{2017}\natexlab{}.
\newblock \showarticletitle{Evolution strategies as a scalable alternative to
  reinforcement learning}.
\newblock \bibinfo{journal}{\emph{arXiv preprint arXiv:1703.03864}}
  (\bibinfo{year}{2017}).
\newblock


\bibitem[\protect\citeauthoryear{Scarselli, Gori, Tsoi, Hagenbuchner, and
  Monfardini}{Scarselli et~al\mbox{.}}{2008}]%
        {scarselli2008graph}
\bibfield{author}{\bibinfo{person}{Franco Scarselli}, \bibinfo{person}{Marco
  Gori}, \bibinfo{person}{Ah~Chung Tsoi}, \bibinfo{person}{Markus
  Hagenbuchner}, {and} \bibinfo{person}{Gabriele Monfardini}.}
  \bibinfo{year}{2008}\natexlab{}.
\newblock \showarticletitle{The graph neural network model}.
\newblock \bibinfo{journal}{\emph{IEEE transactions on neural networks}}
  \bibinfo{volume}{20}, \bibinfo{number}{1} (\bibinfo{year}{2008}),
  \bibinfo{pages}{61--80}.
\newblock


\bibitem[\protect\citeauthoryear{Shapiro}{Shapiro}{2019}]%
        {shapiro2019embodied}
\bibfield{author}{\bibinfo{person}{Lawrence Shapiro}.}
  \bibinfo{year}{2019}\natexlab{}.
\newblock \bibinfo{booktitle}{\emph{Embodied cognition}}.
\newblock \bibinfo{publisher}{Routledge}.
\newblock


\bibitem[\protect\citeauthoryear{Siciliano, Khatib, and Kr{\"o}ger}{Siciliano
  et~al\mbox{.}}{2008}]%
        {siciliano2008springer}
\bibfield{author}{\bibinfo{person}{Bruno Siciliano}, \bibinfo{person}{Oussama
  Khatib}, {and} \bibinfo{person}{Torsten Kr{\"o}ger}.}
  \bibinfo{year}{2008}\natexlab{}.
\newblock \bibinfo{booktitle}{\emph{Springer handbook of robotics}}.
  Vol.~\bibinfo{volume}{200}.
\newblock \bibinfo{publisher}{Springer}.
\newblock


\bibitem[\protect\citeauthoryear{Sims}{Sims}{1994}]%
        {sims1994evolving}
\bibfield{author}{\bibinfo{person}{Karl Sims}.}
  \bibinfo{year}{1994}\natexlab{}.
\newblock \showarticletitle{Evolving virtual creatures}. In
  \bibinfo{booktitle}{\emph{Proceedings of the 21st annual conference on
  Computer graphics and interactive techniques}}. ACM, \bibinfo{pages}{15--22}.
\newblock


\bibitem[\protect\citeauthoryear{Such, Madhavan, Conti, Lehman, Stanley, and
  Clune}{Such et~al\mbox{.}}{2017}]%
        {such2017deep}
\bibfield{author}{\bibinfo{person}{Felipe~Petroski Such},
  \bibinfo{person}{Vashisht Madhavan}, \bibinfo{person}{Edoardo Conti},
  \bibinfo{person}{Joel Lehman}, \bibinfo{person}{Kenneth~O Stanley}, {and}
  \bibinfo{person}{Jeff Clune}.} \bibinfo{year}{2017}\natexlab{}.
\newblock \showarticletitle{Deep neuroevolution: Genetic algorithms are a
  competitive alternative for training deep neural networks for reinforcement
  learning}.
\newblock \bibinfo{journal}{\emph{arXiv preprint arXiv:1712.06567}}
  (\bibinfo{year}{2017}).
\newblock


\bibitem[\protect\citeauthoryear{Sudhakaran, Grbic, Li, Katona, Najarro,
  Glanois, and Risi}{Sudhakaran et~al\mbox{.}}{2021}]%
        {sudhakaran2021growing}
\bibfield{author}{\bibinfo{person}{Shyam Sudhakaran}, \bibinfo{person}{Djordje
  Grbic}, \bibinfo{person}{Siyan Li}, \bibinfo{person}{Adam Katona},
  \bibinfo{person}{Elias Najarro}, \bibinfo{person}{Claire Glanois}, {and}
  \bibinfo{person}{Sebastian Risi}.} \bibinfo{year}{2021}\natexlab{}.
\newblock \showarticletitle{Growing 3D Artefacts and Functional Machines with
  Neural Cellular Automata}.
\newblock \bibinfo{journal}{\emph{arXiv preprint arXiv:2103.08737}}
  (\bibinfo{year}{2021}).
\newblock


\bibitem[\protect\citeauthoryear{Sui, Cai, Bie, Zhang, Zhao, and Zhu}{Sui
  et~al\mbox{.}}{2020}]%
        {sui2020automatic}
\bibfield{author}{\bibinfo{person}{Xin Sui}, \bibinfo{person}{Hegao Cai},
  \bibinfo{person}{Dongyang Bie}, \bibinfo{person}{Yu Zhang},
  \bibinfo{person}{Jie Zhao}, {and} \bibinfo{person}{Yanhe Zhu}.}
  \bibinfo{year}{2020}\natexlab{}.
\newblock \showarticletitle{Automatic generation of locomotion patterns for
  soft modular reconfigurable robots}.
\newblock \bibinfo{journal}{\emph{Applied Sciences}} \bibinfo{volume}{10},
  \bibinfo{number}{1} (\bibinfo{year}{2020}), \bibinfo{pages}{294}.
\newblock


\bibitem[\protect\citeauthoryear{Talamini, Medvet, Bartoli, and
  De~Lorenzo}{Talamini et~al\mbox{.}}{2019}]%
        {talamini2019evolutionary}
\bibfield{author}{\bibinfo{person}{Jacopo Talamini}, \bibinfo{person}{Eric
  Medvet}, \bibinfo{person}{Alberto Bartoli}, {and} \bibinfo{person}{Andrea
  De~Lorenzo}.} \bibinfo{year}{2019}\natexlab{}.
\newblock \showarticletitle{Evolutionary Synthesis of Sensing Controllers for
  Voxel-based Soft Robots}. In \bibinfo{booktitle}{\emph{Artificial Life
  Conference Proceedings}}. MIT Press, \bibinfo{pages}{574--581}.
\newblock


\bibitem[\protect\citeauthoryear{Tang and Ha}{Tang and Ha}{2021}]%
        {tang2021sensory}
\bibfield{author}{\bibinfo{person}{Yujin Tang} {and} \bibinfo{person}{David
  Ha}.} \bibinfo{year}{2021}\natexlab{}.
\newblock \showarticletitle{The sensory neuron as a transformer:
  Permutation-invariant neural networks for reinforcement learning}.
\newblock \bibinfo{journal}{\emph{Advances in Neural Information Processing
  Systems}}  \bibinfo{volume}{34} (\bibinfo{year}{2021}).
\newblock


\bibitem[\protect\citeauthoryear{Tang, Nguyen, and Ha}{Tang
  et~al\mbox{.}}{2020a}]%
        {tang2020neuroevolution}
\bibfield{author}{\bibinfo{person}{Yujin Tang}, \bibinfo{person}{Duong Nguyen},
  {and} \bibinfo{person}{David Ha}.} \bibinfo{year}{2020}\natexlab{a}.
\newblock \showarticletitle{Neuroevolution of self-interpretable agents}. In
  \bibinfo{booktitle}{\emph{Proceedings of the 2020 Genetic and Evolutionary
  Computation Conference}}. \bibinfo{pages}{414--424}.
\newblock


\bibitem[\protect\citeauthoryear{Tang, Tan, and Harada}{Tang
  et~al\mbox{.}}{2020b}]%
        {tang2020learning}
\bibfield{author}{\bibinfo{person}{Yujin Tang}, \bibinfo{person}{Jie Tan},
  {and} \bibinfo{person}{Tatsuya Harada}.} \bibinfo{year}{2020}\natexlab{b}.
\newblock \showarticletitle{Learning agile locomotion via adversarial
  training}. In \bibinfo{booktitle}{\emph{2020 IEEE/RSJ International
  Conference on Intelligent Robots and Systems (IROS)}}. IEEE,
  \bibinfo{pages}{6098--6105}.
\newblock


\bibitem[\protect\citeauthoryear{Tang, Tian, and Ha}{Tang
  et~al\mbox{.}}{2022}]%
        {evojax2022}
\bibfield{author}{\bibinfo{person}{Yujin Tang}, \bibinfo{person}{Yingtao Tian},
  {and} \bibinfo{person}{David Ha}.} \bibinfo{year}{2022}\natexlab{}.
\newblock \showarticletitle{EvoJAX: Hardware-Accelerated Neuroevolution}.
\newblock \bibinfo{journal}{\emph{arXiv preprint arXiv:2202.05008}}
  (\bibinfo{year}{2022}).
\newblock


\bibitem[\protect\citeauthoryear{Treisman, Vieira, and Hayes}{Treisman
  et~al\mbox{.}}{1992}]%
        {treisman1992automaticity}
\bibfield{author}{\bibinfo{person}{Anne Treisman}, \bibinfo{person}{Alfred
  Vieira}, {and} \bibinfo{person}{Amy Hayes}.} \bibinfo{year}{1992}\natexlab{}.
\newblock \showarticletitle{Automaticity and preattentive processing}.
\newblock \bibinfo{journal}{\emph{The American journal of psychology}}
  (\bibinfo{year}{1992}), \bibinfo{pages}{341--362}.
\newblock


\bibitem[\protect\citeauthoryear{Vaswani, Shazeer, Parmar, Uszkoreit, Jones,
  Gomez, Kaiser, and Polosukhin}{Vaswani et~al\mbox{.}}{2017}]%
        {vaswani2017attention}
\bibfield{author}{\bibinfo{person}{Ashish Vaswani}, \bibinfo{person}{Noam
  Shazeer}, \bibinfo{person}{Niki Parmar}, \bibinfo{person}{Jakob Uszkoreit},
  \bibinfo{person}{Llion Jones}, \bibinfo{person}{Aidan~N Gomez},
  \bibinfo{person}{{\L}ukasz Kaiser}, {and} \bibinfo{person}{Illia
  Polosukhin}.} \bibinfo{year}{2017}\natexlab{}.
\newblock \showarticletitle{Attention is all you need}. In
  \bibinfo{booktitle}{\emph{Advances in neural information processing
  systems}}. \bibinfo{pages}{5998--6008}.
\newblock


\bibitem[\protect\citeauthoryear{Wang, Liao, Ba, and Fidler}{Wang
  et~al\mbox{.}}{2018}]%
        {wang2018nervenet}
\bibfield{author}{\bibinfo{person}{Tingwu Wang}, \bibinfo{person}{Renjie Liao},
  \bibinfo{person}{Jimmy Ba}, {and} \bibinfo{person}{Sanja Fidler}.}
  \bibinfo{year}{2018}\natexlab{}.
\newblock \showarticletitle{Nervenet: Learning structured policy with graph
  neural networks}. In \bibinfo{booktitle}{\emph{International Conference on
  Learning Representations}}.
\newblock


\bibitem[\protect\citeauthoryear{Wong, B{\"a}ck, Kononova, and Plaat}{Wong
  et~al\mbox{.}}{2021}]%
        {wong2021multiagent}
\bibfield{author}{\bibinfo{person}{Annie Wong}, \bibinfo{person}{Thomas
  B{\"a}ck}, \bibinfo{person}{Anna~V Kononova}, {and} \bibinfo{person}{Aske
  Plaat}.} \bibinfo{year}{2021}\natexlab{}.
\newblock \showarticletitle{Multiagent deep reinforcement learning: Challenges
  and directions towards human-like approaches}.
\newblock \bibinfo{journal}{\emph{arXiv preprint arXiv:2106.15691}}
  (\bibinfo{year}{2021}).
\newblock


\bibitem[\protect\citeauthoryear{Wu, Xu, Dai, Wan, Zhang, Yan, Tomizuka,
  Gonzalez, Keutzer, and Vajda}{Wu et~al\mbox{.}}{2020}]%
        {wu2020visual}
\bibfield{author}{\bibinfo{person}{Bichen Wu}, \bibinfo{person}{Chenfeng Xu},
  \bibinfo{person}{Xiaoliang Dai}, \bibinfo{person}{Alvin Wan},
  \bibinfo{person}{Peizhao Zhang}, \bibinfo{person}{Zhicheng Yan},
  \bibinfo{person}{Masayoshi Tomizuka}, \bibinfo{person}{Joseph Gonzalez},
  \bibinfo{person}{Kurt Keutzer}, {and} \bibinfo{person}{Peter Vajda}.}
  \bibinfo{year}{2020}\natexlab{}.
\newblock \showarticletitle{Visual transformers: Token-based image
  representation and processing for computer vision}.
\newblock \bibinfo{journal}{\emph{arXiv preprint arXiv:2006.03677}}
  (\bibinfo{year}{2020}).
\newblock


\bibitem[\protect\citeauthoryear{Yim, Shen, Salemi, Rus, Moll, Lipson, Klavins,
  and Chirikjian}{Yim et~al\mbox{.}}{2007}]%
        {yim2007modular}
\bibfield{author}{\bibinfo{person}{Mark Yim}, \bibinfo{person}{Wei-Min Shen},
  \bibinfo{person}{Behnam Salemi}, \bibinfo{person}{Daniela Rus},
  \bibinfo{person}{Mark Moll}, \bibinfo{person}{Hod Lipson},
  \bibinfo{person}{Eric Klavins}, {and} \bibinfo{person}{Gregory~S
  Chirikjian}.} \bibinfo{year}{2007}\natexlab{}.
\newblock \showarticletitle{Modular self-reconfigurable robot systems [grand
  challenges of robotics]}.
\newblock \bibinfo{journal}{\emph{IEEE Robotics \& Automation Magazine}}
  \bibinfo{volume}{14}, \bibinfo{number}{1} (\bibinfo{year}{2007}),
  \bibinfo{pages}{43--52}.
\newblock


\bibitem[\protect\citeauthoryear{Zambaldi, Raposo, Santoro, Bapst, Li,
  Babuschkin, Tuyls, Reichert, Lillicrap, Lockhart, et~al\mbox{.}}{Zambaldi
  et~al\mbox{.}}{2018}]%
        {zambaldi2018deep}
\bibfield{author}{\bibinfo{person}{Vinicius Zambaldi}, \bibinfo{person}{David
  Raposo}, \bibinfo{person}{Adam Santoro}, \bibinfo{person}{Victor Bapst},
  \bibinfo{person}{Yujia Li}, \bibinfo{person}{Igor Babuschkin},
  \bibinfo{person}{Karl Tuyls}, \bibinfo{person}{David Reichert},
  \bibinfo{person}{Timothy Lillicrap}, \bibinfo{person}{Edward Lockhart},
  {et~al\mbox{.}}} \bibinfo{year}{2018}\natexlab{}.
\newblock \showarticletitle{Deep reinforcement learning with relational
  inductive biases}. In \bibinfo{booktitle}{\emph{International Conference on
  Learning Representations}}.
\newblock


\end{thebibliography}
